\documentclass{article} 
\usepackage{iclr2027/iclr2027_conference,times}  

\usepackage{amsmath,amsfonts,bm}

\def\eqref#1{equation~\ref{#1}}

\def\1{\bm{1}}

\def\vk{{\bm{k}}}

\def\vo{{\bm{o}}}

\def\vq{{\bm{q}}}

\def\vu{{\bm{u}}}
\def\vv{{\bm{v}}}

\def\vy{{\bm{y}}}

\def\mA{{\bm{A}}}

\def\mI{{\bm{I}}}

\def\mQ{{\bm{Q}}}
\def\mR{{\bm{R}}}
\def\mS{{\bm{S}}}

\def\mX{{\bm{X}}}
\def\mY{{\bm{Y}}}

\DeclareMathAlphabet{\mathsfit}{\encodingdefault}{\sfdefault}{m}{sl}
\SetMathAlphabet{\mathsfit}{bold}{\encodingdefault}{\sfdefault}{bx}{n}

\newcommand{\R}{\mathbb{R}}

\usepackage{hyperref}
\usepackage{url}
\usepackage{booktabs}
\usepackage{caption}
\usepackage{subcaption}
\usepackage{graphicx}
\usepackage{multirow}
\usepackage[table,xcdraw]{xcolor}
\usepackage{amsmath}
\usepackage{amsthm}
\usepackage{amssymb}
\usepackage{amsfonts}
\usepackage{stmaryrd}
\usepackage{enumitem}
\usepackage{algorithm}
\usepackage{algpseudocode}
\usepackage{fontawesome5}

\usepackage{adjustbox}
\usepackage{tabularx}

\newcolumntype{Y}{>{\centering\arraybackslash}X}

\definecolor{candypink}{rgb}{1.0, 0.41, 0.71}
\definecolor{metablue}{rgb}{0.2, 0.4, 0.8}
\definecolor{phthaloblue}{rgb}{0.0, 0.15, 0.45}
\definecolor{green(ncs)}{rgb}{0.0, 0.62, 0.42}
\definecolor{lightwheat}{rgb}{1.0, 0.98, 0.82}
\definecolor{rosepink}{HTML}{F4A7B9}
\definecolor{skyblue}{HTML}{A7C7E7}
\definecolor{palatinateblue}{rgb}{0.15, 0.23, 0.89}

\usepackage{listings}
\usepackage{xcolor}
\usepackage{titletoc}

\usepackage[T1]{fontenc}  
\definecolor{jsaffil}{HTML}{6E6466}  
\makeatletter
\newcommand\authornote[2]{%
  \textsuperscript{\ensuremath{#1}}%
  \if\relax\detokenize{#2}\relax\else
    \g@addto@macro\@thanks{\iclr@authorfootnote{#1}{#2}}%
  \fi}
\long\def\iclr@authorfootnote#1#2{%
  \begingroup\def\@thefnmark{#1}\@footnotetext{#2}\endgroup}
\renewcommand\and{\end{tabular}%
  \hskip 0pt plus 3em\penalty0\hskip 0pt plus -3em
  \begin{tabular}[t]{l}\bf\ignorespaces}
\newcommand\shorttitle[1]{\lhead{#1}}
\def\iclr@affiliation{}
\newcommand\affiliation[1]{\gdef\iclr@affiliation{#1}}
\AtBeginDocument{\ifx\iclr@affiliation\@empty\else
  \g@addto@macro\@author{\end{tabular}\par
    \begin{tabular}[t]{l}\iclr@affiliation}\fi}
\makeatother

\title{History Replays Itself: Unleashing Long-Context Capability in Linear Attention}

\author{Li Nanbo\authornote{*}{Equal contribution.} \quad
    Firas Laakom\authornote{*}{} \quad
    Yilan Zhang \quad
    Piotr Pi\k{e}kos \quad
    Yimeng Chen \\
    \textbf{Liangyu Wang} \quad
    \textbf{Yifan Xu} \quad
    \textbf{Mateusz Ostaszewski} \quad
    \textbf{Dandan Guo} \quad
    \textbf{Wenyi Wang} \\
    \textbf{Yanning Dai} \quad
    \textbf{Yuhui Wang} \quad
    \textbf{J\"{u}rgen Schmidhuber} \vspace{0.5em}\\ 
    King Abdullah University of Science and Technology (KAUST) \\
    \faGithub~\textcolor{palatinateblue}{\url{https://nanboli.github.io/notebank-page}}
}

\shorttitle{NoteBank: History Replays Itself} 

\newif\ifshowtmp
\showtmptrue

\ifshowtmp

\newcommand{\nanbo}[1]{\textcolor{blue}{({Nanbo:} #1)}}
\newcommand{\wenyi}[1]{\textcolor{pink}{({Wenyi:} #1)}} 
\newcommand{\yimeng}[1]{\textcolor{orange}{{Yimeng:} #1)}}
\newcommand{\piotr}[1]{\textcolor{red}{({Piotr:} #1)}}
\newcommand{\firas}[1]{\textcolor{red}{({Firas:} #1)}}
\newcommand{\yifan}[1]{\textcolor{cyan}{({Yifan:} #1)}}
\newcommand{\yuhui}[1]{\textcolor{teal}{({Yuhui}: #1)}}
\newcommand{\yanning}[1]{\textcolor{blue}{({Yanning}: #1)}}

\else

\newcommand{\nanbo}[1]{}
\newcommand{\wenyi}[1]{}
\newcommand{\yimeng}[1]{}
\newcommand{\piotr}[1]{}
\newcommand{\firas}[1]{}
\newcommand{\yifan}[1]{}
\newcommand{\yanning}[1]{}
\newcommand{\yuhui}[1]{}
\newcommand{\juergen}[1]{}

\fi

\iclrfinalcopy 

\begin{document}

\maketitle

\begin{abstract}
Linear attention promises the efficiency needed for long-context language modelling, yet its fixed-size recurrent state must simultaneously support ongoing computation and preserve information from an ever-growing history. This coupling has reinforced a prevailing view that the state's memory capacity defines the model's long-context capability. We challenge this presumed ceiling: \emph{recurrent computation need not bear the full burden of long-term storage}. We introduce NoteBank, a sparse episodic memory that archives selected text before future queries are known, uses the backbone's native features for writing and addressing, and reads retrieved evidence through its native forward computation via replay. This \emph{training-free, plug-and-play} interface requires no external embedding model, separately trained retriever, or separate neural reader. At one million tokens, it raises Gated DeltaNet's mean accuracy across six RULER retrieval tasks from 1.7\% to 70.3\%, with no observed significant retrieval degradation from 4K to 1M tokens. This recovery adds $\leqslant$0.025\% to analytical FLOPs, with persistent context storage approximately 1/2{,}600 that of a Transformer's full MHA KV cache at matched width, depth, and precision. Experiments extend from three 1.3B recurrent backbones to Mamba-2-8B and post-trained Qwen3.5-9B, demonstrating gains on LongBench v1/v2 and MemoryAgentBench. Together, these results show that sparse evidence storage and native replay can substantially expand the long-context capability of pretrained linear attention models while retaining the efficiency of recurrent computation.
\end{abstract}

\begin{figure}[h!]
    \begin{center}
        \includegraphics[width=\linewidth]{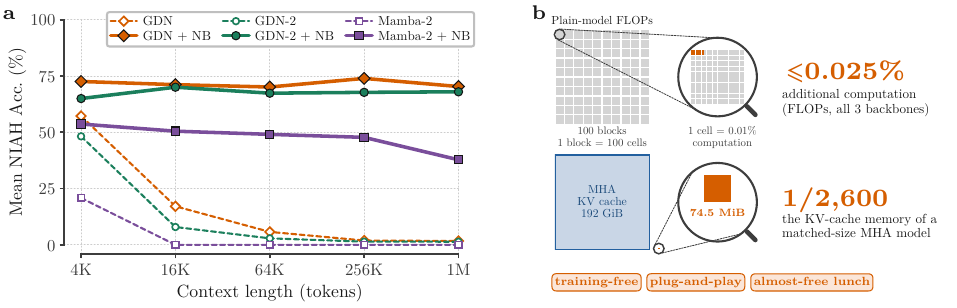}
    \end{center}
    \caption{\textbf{NoteBank (NB) recovers million-token retrieval with negligible compute overhead}. \textbf{a}, Mean accuracy over six RULER-NIAH retrieval tasks for three pretrained 1.3B recurrent models, without NoteBank (dashed) and with training-free top-2 note replay (solid). \textbf{b}, At 1M tokens, NoteBank adds $\leqslant$0.025\% layer-level analytical FLOPs across the three backbones (GDN shown). GDN + NoteBank requires 74.5 MiB of persistent context storage, including its recurrent state---approximately $1/2{,}600$ of the 192 GiB MHA KV cache at matched width, depth, and \texttt{BF16} precision. Square areas represent storage, with a magnified inset for NoteBank. FLOPs include all evaluated stages at the full-model level and matched output length; they do not measure runtime. Storage excludes model weights and transient workspace.}
    \label{fig:intro_demo}
\end{figure}

\section{Introduction}\label{sec:intro}
Long-context capability enables language models to reason over extensive documents, conversations, codebases, and interaction histories. Yet making these histories available to a model is costly: full softmax attention~\citep{vaswani2017attention,shazeer2019fasttransformerdecoding} requires quadratic attention computation during context processing and a key–value cache that grows linearly with sequence length. Linear attention~\citep{schmidhuber1992fwp,katharopoulos2020lattn,schlag2021linear} offers a different way to carry information forward. Rather than retaining a separate representation for every past token, it compresses history into a fixed-size memory state that is updated as each new token arrives. This compact representation enables linear-time sequence processing with a constant-size attention state~\citep{sun2023retentive-retnet,peng2023rwkv,yang2023gated-linear-attention,qin2024hgrn2}. Its practical relevance is reflected in recent large-scale hybrid models such as Griffin~\citep{de2024griffin}, Jamba~\citep{lieber2024jamba}, Nemotron-H~\citep{nemotron_h}, MiniMax-01~\citep{li2025minimax01}, Qwen3-Next and Qwen3.5~\citep{qwen3-next,qwen35,qwen36}, and Kimi Linear~\citep{kimiteam2025kimilinear,kimi_k26}, which combine recurrent linear-attention layers with softmax attention to improve inference efficiency. 

This efficiency, however, comes from asking a single recurrent state to serve two roles: ongoing computation and long-term memory~\citep{schmidhuber1992learning,hochreiter1997long,wang2025test-time-regression}. These roles impose competing demands as context grows: computation requires the state to continually transform and compress history, whereas memory requires specific past evidence to remain recoverable even when its relevance emerges only much later. As history grows while the state remains fixed in size, continual compression risks irreversibly discarding evidence before its future relevance is known~\citep{gu2024mamba,glm5_2026}. Prior work establishes finite-state limits on exact copying~\citep{jelassi24repeat},
identifies recency bias and over-smoothing in selective state-space models~\citep{wang2025understanding}, and documents substantial associative-recall gaps in several efficient alternatives to attention~\citep{arora24tradeoff}. Improved state-update rules~\citep{schlag2021linear,yang2024deltanet,yang2025gateddeltanet,hatamizadeh2026gateddeltanet2,kimiteam2025kimilinear,peng2025rwkv7gooseexpressivedynamic,hu2025comba,siems2025deltaproduct,sun2024tttlayers,behrouz2025titans,von2025mesanet} can manage this limited capacity more effectively, but retain the same underlying constraint: the bounded state must still support evolving computation while preserving evidence from an ever-growing history. Together, these limitations reinforce a familiar picture: linear attention is efficient, yet its fixed-state memory appears to impose a fundamental ceiling on long-context capability.

We challenge this apparent ceiling: \emph{poor long-range retention through a recurrent state need not imply an inability to use distant evidence.} We introduce NoteBank, a \textbf{training-free, plug-and-play} sparse episodic memory that separates evidence preservation from recurrent computation: selected text spans are archived before any question is observed, addressed using the backbone's native features, and read through its ordinary forward computation via replay. NoteBank augments existing recurrent and hybrid checkpoints without changing backbone weights or architecture, requiring no external embedding model, separately trained retriever, or separate neural reader. 

As Figure~\ref{fig:intro_demo} shows, NoteBank raises Gated DeltaNet's mean accuracy across six RULER~\citep{hsieh2024ruler} retrieval tasks at one million tokens from 1.7\% to 70.3\%, with \textbf{negligible added computation}: less than 0.026\% in analytical layer-level FLOPs. At the same length, GDN + NoteBank requires approximately 1/2600 of the persistent context storage of a full MHA KV cache at matched width, depth, and precision. Beyond controlled retrieval, its benefits extend from 1.3B research models to post-trained Mamba-2-8B~\citep{dao2024transformers} and Qwen3.5-9B~\citep{qwen35}, with improvements on LongBench-v1/v2~\citep{bai2024longbench,bai2025longbenchv2} and MemoryAgentBench~\citep{hu2025memoryagentbench}. These results show that the long-context capability of pretrained recurrent models can substantially exceed what their fixed-state memory interface alone exposes.

Our contributions are threefold:
\begin{itemize}
    \setlength{\itemsep}{0pt}
    \setlength{\parsep}{0pt}
    \item \textbf{Breaking the apparent fixed-state ceiling.} 
    We show that the long-context capability of pretrained recurrent models can substantially exceed what their conventional fixed-state memory interface exposes: poor long-range retention through the recurrent state need not imply an inability to use distant evidence once that evidence is independently preserved and made accessible.

    \item \textbf{Turning pretrained computation into episodic memory.}
    NoteBank decouples evidence preservation from recurrent computation through question-independent sparse text archiving, native-feature addressing, and on-demand replay. This \textbf{\textit{training-free, plug-and-play}} interface uses the backbone itself to write, address, and read memory, requiring neither retraining nor an external embedding model or separate neural reader.

    \item \textbf{Recovering capability at \textit{near-zero} compute overhead.}
    Across recurrent and hybrid backbones, NoteBank recovers strong performance from controlled million-token retrieval to natural-document and agent-memory tasks, while adding negligible analytical computation and requiring orders of magnitude less persistent context storage than a matched full-attention KV cache.
\end{itemize}

\section{Preliminary}\label{sec:pre}

\textbf{History compression with linear attention.}
A broad family of linear-attention models compresses the processed history into recurrent key-value states~\citep{schmidhuber1992fwp,katharopoulos2020lattn,schlag2021linear}. For a single layer and head, the backbone's existing feature transformations produce query and key vectors $\vq_t,\vk_t\in\R^{d_k}$ and a value vector $\vv_t\in\R^{d_v}$ from its input representations at token position $t$. The key-value pair updates the recurrent memory, while the query reads from it~\citep{yang2024deltanet,dao2024transformers}: 
\begin{equation}
    \mS_t = \mS_{t-1}\mA_t + \beta_t\vv_t\vk_t^{\top},
    \qquad
    \vo_t = \mS_t\vq_t,
\label{eq:lattn}
\end{equation}
where $\mS_t\in\R^{d_v\times d_k}$ serves as a compressed memory of the prefix, $\mA_t\in\R^{d_k\times d_k}$ controls memory retention and transformation, and $\beta_t$ controls the write strength. Standard additive linear attention~\citep{Schmidhuber:91fastweights,katharopoulos2020lattn} sets $\mA_t=\mI$ and $\beta_t=1$; gated~\citep{sun2023retentive-retnet,yang2023gated-linear-attention,gu2024mamba,dao2024transformers,beck2024xlstm} and delta-rule~\citep{schlag2021linear,yang2024deltanet,yang2025gateddeltanet,hatamizadeh2026gateddeltanet2,kimiteam2025kimilinear} variants modify these updates. 
Input scaling is absorbed into this parametrisation; output gates and normalisation are omitted. The same fixed-size state supports both ongoing computation and access to past information: each query reads the history only insofar as it remains encoded in $\mS_t$.

\textbf{Problem setup.} 
We consider pretrained recurrent and hybrid language models that process a context before receiving a question. Let $\mathcal{M}_\phi$ be such a model, $\mX=(x_1,\ldots,x_T)$ the context, and $\mQ=(q_1,\ldots,q_{|\mQ|})$ a subsequent question or prompt. The model reads $\mX$ causally and retains an inference state $\mathcal{S}_T$, consisting of its recurrent states and any auxiliary caches. When $\mQ$ arrives, it generates a response $\mY\sim p_\phi(\cdot\mid\mathcal{S}_T,\mQ)$. Access to the preceding context is therefore limited to what remains in $\mathcal{S}_T$: evidence must be preserved before the model knows which information the question will require. 
\section{NoteBank} \label{sec:method}

\subsection{Decoupling Recurrent Computation from Archival Memory}\label{sec:method:decouple}

NoteBank maintains a sparse episodic archive alongside the backbone's unchanged state updates during context processing. Before observing $\mQ$, it uses native features to select spans, i.e., contiguous text spans (or interchangeably ``segments'') from the context, and index them for later retrieval. When $\mQ$ arrives, its native query features retrieve notes, whose text is replayed through the backbone to support generation. After processing $t$ context tokens, the archive is
\begin{equation}
    \mathcal{NB}_t
      = \{(\mathbf z_i,\mathcal{A}_i)\}_{i=1}^{N_t},
    \qquad
    \mathcal{A}_i
      = \{\bm{\rho}_{i}^{\ell,h}\}_{
          (\ell,h)\in\mathcal L\times\mathcal H},
    \label{eq:nb-bank}
\end{equation}
where $N_t$ is the number of stored notes, $\mathbf z_i$ contains a span's original token IDs, and each address $\bm{\rho}_{i}^{\ell,h}$ is a compact vector derived from native query features within that span. The sets $\mathcal L$ and $\mathcal H$ index recurrent layers and heads; address construction and matching are detailed in Sec.~\ref{sec:method:interface}. Archived text persists independently of subsequent state updates. 
\begin{figure}[t]
    \begin{center}
        \includegraphics[width=\linewidth]{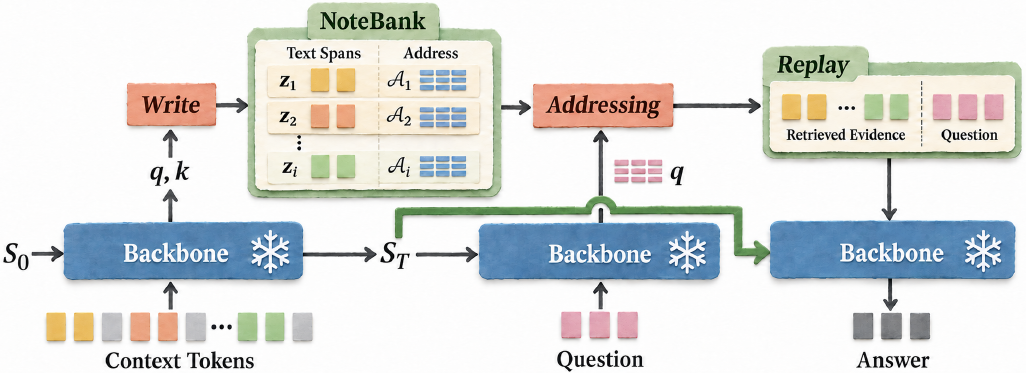}
    \end{center}
    \caption{\textbf{Overview of NoteBank.} During context encoding, native features ``keys'' and ``queries'' (i.e., $\vk$ and $\vq$) guide span selection and indexing; each span $\mathbf{z}_i$ is stored with addresses $\mathcal{A}_i$. Taking the retained context state $\mS_T$, the backbone encodes the question to obtain queries for retrieving relevant spans. The native forward pass starts from the same $\mS_T$ and processes the retrieved evidence and then the question, and generates the answer. Backbone parameters remain frozen throughout. We refer to Algorithm~\ref{alg:notebank} for more details.}
    \label{fig:method}
\end{figure}

\subsection{A Native Memory Interface} \label{sec:method:interface}

\textbf{Write (top-$W$).}
We partition the context into consecutive, non-overlapping chunks of $C$ tokens, carrying the recurrent state across chunk boundaries~\citep{sun2023retentive-retnet,yang2023gated-linear-attention,yang2024fla}. Once a chunk accumulates $C$ tokens, the writer scores its tokens by native-key distinctiveness to select up to $W$ notes:
\begin{equation}
 d_t^{\ell,h}
   = 1-\left\langle
       \widehat{\vk}_t^{\ell,h},
       \overline{\bm{\mu}}_{\mathcal C(t)}^{\ell,h}
     \right\rangle,
 \qquad
 d_t
   = \frac{1}{|\mathcal L\times\mathcal H|}
     \sum_{(\ell,h)\in\mathcal L\times\mathcal H}
       d_t^{\ell,h}.
 \label{eq:nb-write}
\end{equation}
Here $\mathcal{C}(t)$ contains the positions in the chunk containing $t$, and $\overline{\bm{\mu}}_{\mathcal C(t)}^{\ell,h}$ is the normalised mean of the keys in that chunk. Hats $\widehat{\cdot}$ denote $\ell_2$ normalisation throughout this paper. 

To form each note, we score candidate windows, i.e., short contiguous sequences of tokens, by summing $d_t$ over their positions. We choose the highest-scoring window and use the sentence containing its central token as the starting segment, then expand it with adjacent sentences subject to the $B$-token limit. Any sentence overlapping the selected segment is excluded from subsequent selections. For each selected text segment with token positions $I_i$, we store its original token IDs $\mathbf{z}_i$ and construct one address per layer-head channel. Each channel selects the token with the highest key-distinctiveness score and stores a compact version of its native query vector:
\begin{equation}
 t_i^{*,\ell,h}
   = \operatorname*{arg\,max}_{t\in I_i} d_t^{\ell,h},
 \qquad
 \bm{\rho}_i^{\ell,h}
   = \operatorname{compact}
       \left(\vq_{t_i^{*,\ell,h}}^{\ell,h}\right)
     \in\R^{d_\rho}.
 \label{eq:nb-address}
\end{equation}
Thus, each channel indexes the segment using the query at its most distinctive key position; different channels may select different positions. The fixed mapping $\operatorname{compact}$ is a non-trainable feature map for constructing compact addresses; setting $\operatorname{compact}$ to identity presents full native-query addresses. Full details of address construction is provided in App.~\ref{app:address-construction}.

\textbf{Addressing (top-$R$).}
When $\mQ$ arrives, we process it from a copy of the retained inference state $\mathcal{S}_T$ to obtain native query features. For each question position $\tau\in\{1,\ldots,|\mQ|\}$, we construct
$\vu_\tau^{\ell,h}=\operatorname{compact}(\vq_\tau^{\ell,h})$
using the same mapping as the stored addresses. Each layer-head channel scores every stored note $i\in\{1,\ldots,N_T\}$ by the maximum cosine similarity between its address and the question's query features~\citep{khattab2020colbert}, then votes for its highest-scoring note:
\begin{equation}
 s_i^{\ell,h}
   = \max_{1\leq\tau\leq|\mQ|}
       \left\langle
         \widehat{\vu}_\tau^{\ell,h},
         \widehat{\bm{\rho}}_i^{\ell,h}
       \right\rangle,
 \qquad
 V_i
   = \sum_{(\ell,h)\in\mathcal L\times\mathcal H}
       \mathbf{1}\!\left[
         i=\operatorname*{arg\,max}_{j}
         s_j^{\ell,h}
       \right].
 \label{eq:nb-route}
\end{equation}
We retrieve up to $R$ notes with the most votes, with a preference of earlier-inserted notes when vote counts tie. We then concatenate their token IDs in original document order, determined by each note's starting token position, to form the replay sequence $\mR$. The temporary state used to process $\mQ$ is discarded after retrieval.

\textbf{Readout with native replay.}
After addressing, the backbone reads the retrieved text through its native forward computation. Let $\mathcal{S}_{\mathrm{read}}$ be a fresh copy of the retained state $\mathcal{S}_T$, and $F_{\phi}(\mathcal{S},\mX_{\ast})$ denote the backbone's state update when processing a token sequence $\mX_{\ast}$. We replay $\mR$, then process $\mQ$ again to generate the response:
\begin{equation}
 \mathcal{S}_{\mR}
   = F_{\phi}(\mathcal{S}_{\mathrm{read}},\mR),
 \qquad
 \mY \sim p_\phi(\cdot\mid\mathcal{S}_{\mR},\mQ).
 \label{eq:nb-replay}
\end{equation}
Replay introduces at most $R\times B$ tokens and traverses the full backbone, including attention blocks under the configured policy in hybrid models. This readout requires neither additional training nor a separate neural reader. Because it follows native forward computation, backbone-supported parallel algorithms~\citep{yang2024fla} apply directly, without NoteBank-specific customisation.

To improve the model's use of replayed evidence on downstream tasks, we optionally train lightweight LoRA adapters~\citep{hu2022lora} while keeping pretrained weights frozen. Adapters are disabled during context processing and retrieval, leaving writing and addressing unchanged. They are enabled throughout replay and subsequent question answering. See App.~\ref{app:readout-adaptation} for more details.

\subsection{Memory and Computational Overhead}
\label{sec:method:complexity}
We analyse NoteBank's additional costs relative to the same backbone and attention configuration.

\textbf{Memory.}
Let $J=|\mathcal L\times\mathcal H|$ be the number of indexed recurrent layer-head channels. Writing at most $W$ notes per completed $C$-token chunk gives:
\begin{equation}
 N_T \leq W\lfloor T/C\rfloor,
 \qquad
 \mathrm{Storage}_{\mathcal{NB}}
   = O\!\left(N_T(B+Jd_\rho)\right).
\end{equation}
Each note stores at most $B$ token IDs and $J=|\mathcal L\times\mathcal H|$ compact addresses of dimension $d_\rho$. The storage of $\mathcal{NB}$ archive \emph{grows linearly with history length} at a rate controlled by these budgets, without enlarging the recurrent state.

\textbf{Computation.}
Writing reuses native features, adding $O(T)$ work for fixed chunk size and model dimensions. Addressing requires one additional forward pass over the question, followed by exhaustive matching at cost 
$O(|\mQ|N_TJd_\rho)$, 
layer-head channel voting, and top-$R$ selection. Readout adds at most $R \times B$ replay tokens before question processing and generation. Thus, archive storage and exhaustive matching \emph{grow linearly with history length}, while \emph{the replay budget remains constant}; replay computation depends on the backbone and attention policy. 

\section{Recovering Long-Context Capability}\label{sec:recover}
Can pretrained recurrent models recover strong long-context capability when past evidence remains accessible beyond their recurrent states? We investigate this question through long-range retrieval at increasing context lengths, real-world tasks, and controlled restrictions on historical access.

\textbf{Setup.}
We evaluate \textit{publicly available checkpoints} of GDN and GDN-2 (1.3B), Mamba-2 (1.3B and 8B), and Qwen3.5 (9B). For Mamba-2, recurrent-state rescaling is necessary for the model to avoid numerical failure during long-context inference. NB-TF is training-free while NB-LoRA adapts the read stage using lightweight adapters trained on \textit{fewer than 9.3M tokens of general language data}, with one adapter set per backbone shared across downstream tasks. Both variants construct notes \textit{without observing the evaluation question}. Our evaluation covers six RULER-NIAH tasks~\citep{hsieh2024ruler}, LongBench-v1 and v2~\citep{bai2024longbench,bai2025longbenchv2}, and MemoryAgentBench~\citep{hu2025memoryagentbench}. We refer to App.~\ref{app:impl} for more experimental details. 


\begin{figure*}[t]
    \centering
    \begin{minipage}{\textwidth}
        \centering
        \includegraphics[width=\linewidth]{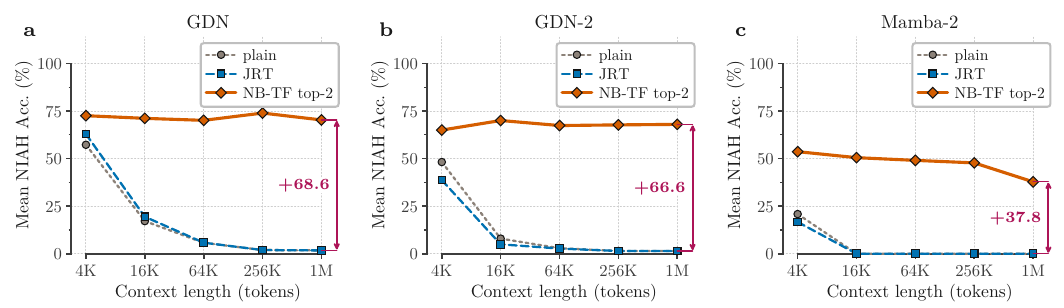}
        \captionof{figure}{\textbf{Recovering long-range retrieval \textit{without training}.} NoteBank (NB-TF top-2) substantially improves retrieval over plain forward pass and full-prompt repetition (JRT), using the same frozen backbone. Scores are averaged over six RULER retrieval tasks, including single- and multi-passage retrieval. See App.~\ref{app:res:complementary} for per-task results.}
        \label{fig:tf_ruler:main}
    \end{minipage}
    \par\vspace{1em}
    \begin{minipage}{\textwidth}
        \centering

        \small
        \renewcommand{\arraystretch}{1.0}
        \setlength{\tabcolsep}{5pt}
        \begin{tabular}{l|l|cccc|c}
            \toprule
            \textbf{Backbone} & \textbf{Method}
            & \textbf{Single-doc QA}
            & \textbf{Multi-doc QA}
            & \textbf{Summarisation}
            & \textbf{Few-shot}
            & \textbf{Overall} \\
            \midrule
            Transformer\textsuperscript{\dag}
            & Plain    & 13.40 & 18.97 & 8.43 & 23.40 & 16.05 \\
            \midrule
            & Plain    & 6.88 & 7.02 & 13.99 & 34.08 & 15.49 \\
            & JRT      & 7.72 & 6.63 & \textbf{14.79} & 33.56 & 15.67 \\
            \rowcolor[HTML]{EAF1F8}\cellcolor{white}
            & \textbf{NB-TF}    & 6.66 & 7.33 & 13.52 & 36.76 & 16.07 \\
            \rowcolor[HTML]{EAF1F8}\cellcolor{white}\multirow{-4}{*}{GDN}
            & \textbf{NB-LoRA} & \textbf{19.06} & \textbf{20.48} & 11.37 & \textbf{41.22} & \textbf{23.03} \\
            \midrule
            & Plain    & 2.22 & 2.10 & 11.27 & 11.56 & 6.79 \\
            & JRT      & 2.39 & 1.89 & 9.21  & 8.34  & 5.12 \\
            \rowcolor[HTML]{EAF1F8}\cellcolor{white}
            & \textbf{NB-TF}    & 5.98 & 5.36 & \textbf{16.51} & 25.88 & 13.43 \\
            \rowcolor[HTML]{EAF1F8}\cellcolor{white}\multirow{-4}{*}{Mamba-2}
            & \textbf{NB-LoRA} & \textbf{17.84} & \textbf{15.54} & 7.46 & \textbf{27.52} & \textbf{17.10} \\
            \bottomrule
        \end{tabular}
        \captionof{table}{\textbf{From retrieval to real-world long-context tasks.} All backbones have 1.3B parameters. Scores are macro-averaged over evaluated LongBench-v1 tasks, within each category and overall. NB-TF requires no additional training; NB-LoRA adapts the read stage. \textsuperscript{\dag}Reference results aggregated from \citet{behrouz2026memorycaching}, whose LongBench models were trained with 16K contexts. Bold marks the best score within each recurrent backbone.}
        \label{tab:longbench:main}
    \end{minipage}
    \par \vspace{1em}
    \begin{minipage}{\textwidth}
        \centering
        \small
        \renewcommand{\arraystretch}{1.0}
        \setlength{\tabcolsep}{14pt}
        \begin{tabular}{l|l|ccr}
            \toprule
            \textbf{Benchmark} & \textbf{Backbone}
            & \textbf{Plain}
            & \textbf{+ NB-TF}
            & \textbf{$\Delta$} \\
            \midrule
            \multirow{2}{*}{MemoryAgentBench} & 
            Mamba-2-8B       & 0.0 & 14.7 & \textbf{+14.7} \\
            & Qwen3.5-9B    & 55.7 & 56.8 & \textbf{+1.1} \\
            \midrule
            \multirow{1}{*}{LongBench-v2} & 
            Qwen3.5-9B    & 46.1 & 49.3 & \textbf{+3.2} \\
            \bottomrule
        \end{tabular}
        \captionof{table}{\textbf{\textit{Plug-and-play} gains on larger models without training.} We report task-macro-averaged scores on the 2,100-question QA subset of MemoryAgentBench and accuracy on the 375-question QA/dialogue subset of LongBench-v2. NB-TF uses top-4 and top-8 retrieval, respectively. All Qwen evaluations use YaRN (official long-context setup). $\Delta$ denotes the gain (\%).
        }
        \label{tab:plug-and-play}
    \end{minipage}
\end{figure*}

\subsection{Training-free long-range retrieval across architectures and lengths}\label{sec:recover:tf}
We first ask: how much long-range retrieval can be recovered by decoupling recurrent computation from archival memory, without updating the model's weights? We compare plain, full-prompt repetition (JRT)~\citep{arora2024justreadtwice}, and NB-TF on GDN, GDN-2, and Mamba-2 using the same checkpoint within each architecture. Evaluation covers three single-needle tasks, \texttt{S-NIAH-1/2/3}, and three \texttt{multi-key}, \texttt{multi-query}, and \texttt{multi-value} tasks, 100 samples per task. Document lengths range from 4K to 1M tokens; JRT processes approximately twice this input length through repetition. NoteBank writes its archive before observing the question.

\textbf{Results \&. Discussion.}
Figure~\ref{fig:tf_ruler:main} reports mean accuracy across the six tasks. Plain and JRT approach zero accuracy as context length grows, whereas NB-TF maintains roughly 70\% accuracy on GDN and GDN-2. Mamba-2 declines more noticeably but retains substantial accuracy at 1M tokens. At 1M, NB-TF exceeds Plain by 68.6, 66.6, and 37.8 percentage points on GDN, GDN-2, and Mamba-2, respectively. The same frozen backbones thus recover substantial retrieval capability through NoteBank's memory, while full-prompt repetition yields little improvement at long contexts. We next test whether these gains translate to real-world tasks.


\subsection{From synthetic retrieval to real-world tasks}\label{sec:recover:realworld}
To test whether NoteBank improves performance beyond synthetic retrieval, and whether read adaptation helps models use the archived evidence, we evaluate GDN and Mamba-2 on LongBench 12 tasks spanning \texttt{single-doc QA}, \texttt{multi-doc QA}, \texttt{summarisation}, and \texttt{few-shot learning}. Within each backbone, we compare Plain, JRT, NB-TF, and NB-LoRA on the same examples. Both NoteBank variants use top-1 replay for GDN and top-4 for Mamba-2; NB-LoRA adapts only the read stage, leaving note writing and addressing unchanged. We also include published Transformer results from \citet{behrouz2026memorycaching} as a reference. Scores are task averages within each category and overall.

\textbf{Results and discussion.}
Table~\ref{tab:longbench:main} shows that NB-TF improves over Plain in six of eight backbone-category comparisons, raising overall scores from 15.49 to 16.07 on GDN and from 6.79 to 13.43 on Mamba-2. GDN with NB-TF reaches an overall score comparable to the Transformer reference (16.05), without additional training. Read adaptation strengthens the gains: NB-LoRA raises overall scores to 23.03 and 17.10, respectively, both above the Transformer reference, with substantial improvements in single- and multi-document QA over the corresponding Plain backbones. 
We hypothesise that replay can introduce distracting evidence, and that read adaptation helps the model use replayed evidence more selectively; we investigate this explanation in Section~\ref{sec:understand}. The benefit of read adaptation remains task-dependent, as NB-LoRA underperforms NB-TF on summarisation.


\subsection{Plug-and-play on off-the-shelf LLMs}\label{sec:recover:plug-and-play}
Can NoteBank improve larger off-the-shelf LLMs, including a strong post-trained hybrid, without additional training? We evaluate Mamba-2-8B and Qwen3.5-9B on selected MemoryAgentBench tasks, and Qwen3.5-9B on the 375-question QA/dialogue subset of LongBench-v2. Within each backbone, we compare Plain and NB-TF using the same checkpoint and examples. NB-TF uses top-4 replay on MemoryAgentBench and top-8 on LongBench-v2. For Qwen, YaRN is enabled for both Plain and NB-TF, following the official recommendation. Retrieval-budget comparisons and task-level results are provided in the App.~\ref{app:res:complementary}.

\textbf{Results and discussion.}
Table~\ref{tab:plug-and-play} shows gains across all three model-benchmark pairs. On MemoryAgentBench, plain Mamba-2 fails completely while, with NB-TF, it improves from 0.0 to 14.7. Qwen improves from 55.7 to 56.8 on MemoryAgentBench and from 46.1 to 49.3 on LongBench v2, demonstrating benefits even for a strong post-trained hybrid with existing long-context capabilities. We attribute Mamba-2’s complete failure in this setting to long-context state instability (detailed in Appendix~\ref{app:res:complementary}). Together, these results support NoteBank as a plug-and-play memory extension across recurrent and hybrid architectures, without additional training.

\section{Almost a Free Lunch?}\label{sec:freelunch}
\begin{figure}[t]
    \begin{center}
        \includegraphics[width=\linewidth]{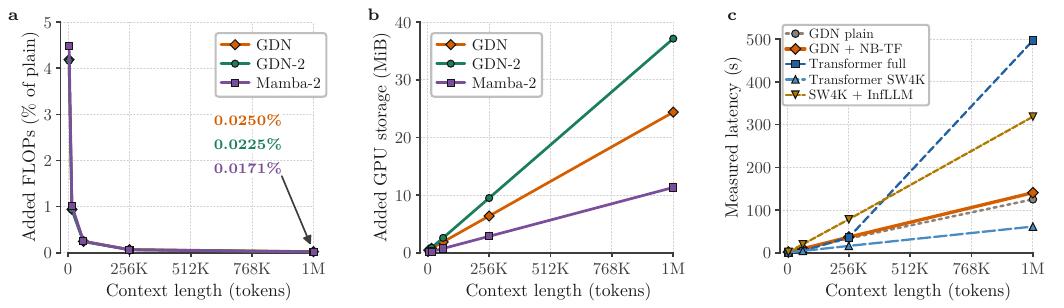}
    \end{center}
    \caption{\textbf{Efficiency of training-free NoteBank.}
        (a) Relative analytical FLOPs overhead, decreasing with context length as fixed costs are amortised.
        (b) Incremental persistent GPU storage grows linearly with context.
        (c) Measured inference latency on an H100, with batch size 1 and 32 generated tokens. NB uses top-2 replay; InfLLM uses the official default top-96.}
    \label{fig:efficiency_scaling}
\end{figure}

Having shown that NoteBank recovers substantial long-context capability, we now ask: at what cost? We quantify its incremental computation and storage analytically, then measure end-to-end latency and peak GPU memory, relating these costs to retrieval quality (Figure~\ref{fig:efficiency_scaling}; Table~\ref{tab:efficiency_1m}).  

\subsection{Negligible marginal computation} \label{sec:freelunch:compute}
We analytically account for Plain and NB-TF inference using the recorded input, archive, and replay lengths, with top-2 replay and matched output lengths. Figure~\ref{fig:efficiency_scaling}a plots the additional FLOPs relative to Plain across context lengths for three recurrent backbones.  The relative overhead decreases as the bounded costs of question encoding and replay are amortised over longer context processing. For fixed question length and replay budget, writing a fixed number of notes per chunk and routing over the resulting archive introduce small linear terms, so the relative overhead approaches a small non-zero floor. At 1M tokens, NoteBank adds only 0.0171\%-0.0250\% to full-model analytical FLOPs, with routing accounting for just 0.50-1.72 GFLOPs. The breakdown in Figure~\ref{fig:efficiency:breakdown} shows nearly constant question-encoding and replay costs, while writing and routing grow linearly but remain negligible relative to backbone computation, even at 1M tokens. Full accounting is provided in App.~\ref{app:implementation-efficiency}.

\subsection{Compact persistent storage}\label{sec:freelunch:storage}
We account for persistent context storage as the backbone's recurrent and convolution states, GPU-resident note addresses and helper tensors, and CPU-resident token IDs and span metadata. Figure~\ref{fig:efficiency_scaling}b shows the additional GPU storage relative to Plain. With a fixed number of notes written per chunk, the address bank grows linearly with context length, while the backbone states and helper tensors remain fixed in size. At 1M tokens, GDN adds 24.4\,MiB of GPU addresses and helper tensors, plus approximately 1.0\,MiB of CPU text and metadata accounted for in compact form. Including backbone states, total persistent context storage is 74.5\,MiB, approximately $1/2600$ of a theoretical \texttt{BF16} full MHA KV cache matched in context length, width, and depth. This accounting excludes model weights, transient allocations, and Python-object overhead (see App.~\ref{app:persistent-memory}).

\subsection{Practical runtime and memory}\label{sec:freelunch:practical}
We measure end-to-end latency and peak GPU memory on an H100 with batch size 1 and 32 generated tokens. For reference, we also evaluate a 1.3B Transformer with full attention, a 4K sliding window (SW4K), and the official InfLLM implementation~\citep{xiao2024infllm}. Figure~\ref{fig:efficiency_scaling}c shows approximately linear latency scaling for GDN with and without NoteBank. Across the matched measurements at 4K, 16K, 64K, and 1M, NoteBank increases latency by 10.6\%-13.2\%. At 1M tokens, latency rises from 125.11 to 140.64\,s,
while peak GPU memory, including model weights, rises from 2.93 to 3.27\,GiB. Measurement protocols and cross-system quality-cost comparisons are provided in App.~\ref{app:implementation-efficiency}. 

Together, these results show that NoteBank extends pretrained linear attention models with substantially stronger long-context retrieval, requiring negligible added arithmetic, compact storage, and modest runtime overhead---\emph{an almost free lunch}.

\begin{figure*}[t]
    \centering

    \begin{minipage}[t]{0.62\textwidth}
        \begin{minipage}[t]{\linewidth}
            \textbf{a \quad Evidence: preserved vs. accessed}
            \vspace{0.4em}
            \centering
            \small
            \includegraphics[width=\linewidth]{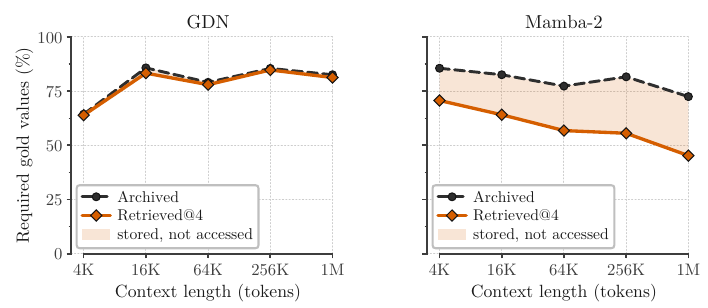}
        \end{minipage}

        \par
        \vspace{0.4em}
        \noindent
        \begin{minipage}[t]{\linewidth}
            \centering
            {\color{jsaffil!30} \rule{0.9\linewidth}{1pt}}
        \end{minipage}

        \par
        \vspace{0.8em}
        \noindent
        \begin{minipage}[t]{\linewidth}
            \centering
            \textbf{c \quad Access intervention on natural-language QA}
            \vspace{0.4em}

            \small
            \setlength{\tabcolsep}{8pt}
            \begin{tabular}{llrrr}
                \toprule
                \textbf{Backbone} & \textbf{Reader}
                & \multicolumn{1}{c}{\textbf{NB}-top1}
                & \multicolumn{1}{c}{\textbf{Oracle}-{1}}
                & \multicolumn{1}{c}{$\boldsymbol{\Delta}$} \\
                \midrule
                \multirow{2}{*}{GDN}
                    & \texttt{Frozen} & 7.05  & 7.59  & \textbf{+0.54}  \\
                    & \texttt{LoRA}   & 19.11 & 22.58 & \textbf{+3.47}  \\
                \midrule
                \multirow{2}{*}{Mamba-2}
                    & \texttt{Frozen} & 4.22  & 7.23  & \textbf{+3.01}  \\
                    & \texttt{LoRA}   & 9.09  & 26.43 & \textbf{+17.34} \\
                \bottomrule
            \end{tabular}
        \end{minipage}
    \end{minipage}
    \hfill
    %
    %
    \begin{minipage}[t]{1pt}
        \vfill
        {\color{jsaffil!30} \rule{1pt}{8cm}}
        \vfill
    \end{minipage}
    \hfill
    \begin{minipage}[t]{0.35\textwidth}
        \centering

        \textbf{b \quad Recoverable performance}
        
        \includegraphics[width=\linewidth]{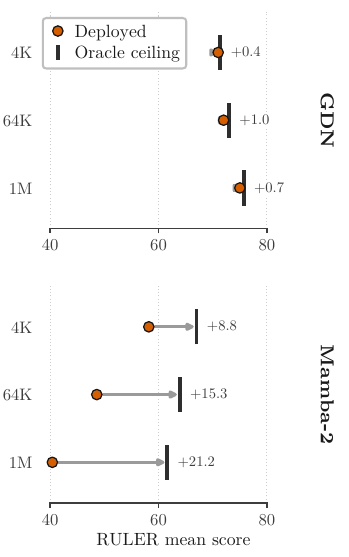}
    \end{minipage}
    \hfill
    \caption{
    \textbf{Evidence preservation, access, and utilisation in NoteBank.}
    \textbf{a}, Gold-value coverage in the archive (\emph{Archived})
    and top-4 retrieved notes (\emph{Retrieved@4});
    shading marks stored but unretrieved values.
    \textbf{b}, Native vs. oracle routing on RULER, with archive, reader, and replay budget fixed.
    \textbf{c}, The same intervention on four LongBench QA tasks (top-1; frozen/LoRA readers).
    Oracle notes contain reference answers, a proxy for
    relevance rather than guaranteed supporting evidence.
    }
    \label{fig:evidence_access}
\end{figure*}

\section{Understanding NoteBank: Evidence Access and Utilisation}\label{sec:understand}
We examine what limits NoteBank's performance by tracing evidence preservation and retrieval, intervening on note selection, and comparing frozen and adapted readers to distinguish the contributions of memory access and read adaptation.

\subsection{What is preserved, and what is retrieved?}\label{sec:understand:access}
Preserving evidence does not guarantee its retrieval. We therefore measure the fractions of required gold values preserved in the archive and recovered by top-4 retrieval, respectively. Figure~\ref{fig:evidence_access}a shows that retrieval closely tracks archival coverage for GDN, whereas Mamba-2 leaves substantially more stored values unretrieved. Holding the archive, reader, and replay budget fixed, oracle note selection raises GDN's score from 75.0 to 75.7 at 1M tokens, but Mamba-2's from 40.4 to 61.6 (Figure~\ref{fig:evidence_access}b). This intervention reveals substantial routing headroom for Mamba-2, with smaller gains for GDN.

We extend this diagnostic to four LongBench QA tasks (\texttt{hotpotqa}, \texttt{2wikimqa}, \texttt{musique}, \texttt{qasper}), using archived spans containing a reference answer as a proxy for relevant evidence. With the archive and reader fixed, we compare native top-1 retrieval with oracle selection of a single answer-bearing note (``Oracle-1'' in Figure~\ref{fig:evidence_access}c). Replaying just this one note instead improves Mamba-2's macro score by 3.01 points with a frozen reader and by 17.34 points with LoRA; GDN gains 0.54 and 3.47 points, respectively. Note that answer containment does not guarantee sufficient supporting evidence, so these results are a proxy diagnostic rather than a performance upper bound. Together, these interventions show that preserved information can remain unaccessed and that the benefit of improved access depends on the reader. 

\subsection{How much does routing matter in practice?}\label{sec:understand:router} 
We next test whether alternative routers can realise this headroom, and at what cost. We hold the archive, reader, and top-1 replay budget fixed and replace only the router. As shown in Table~\ref{tab:router_compr:main}, on RULER at 1M tokens, native routing outperforms random selection, mean-pooled input embeddings, and BGE on both backbones. BM25 achieves higher scores, raising Mamba-2 from 37.6 to 52.7, consistent with the routing headroom identified in Section~\ref{sec:understand:access}. On LongBench with a fixed LoRA reader, differences are modest for GDN but larger for Mamba-2, where BM25 improves the score from 14.25 to 18.66. For GDN at 1M tokens, the default NB router's query encoding and routing take 26\,ms, versus 87\,ms for CPU BM25 in our implementation (App.~\ref{app:res:complementary}, Table~\ref{tab:router_compr:full}). These comparisons show that the default native routing provides effective retrieval with low measured routing latency and no auxiliary encoder, while gains from alternative routers motivate further exploration of the retrieval quality-cost trade-off within NoteBank.

\begin{figure*}[t]
    \centering
    \begin{minipage}{\textwidth}
        \centering
    
        \small
        \renewcommand{\arraystretch}{1.0}
        \setlength{\tabcolsep}{14pt}
        \begin{tabular}{l@{\hspace{12pt}}l!{\hspace{-16pt}}cccc}
            \toprule
            \multirow{2}{*}[-0.8ex]{\textbf{Routing Signal}} & \multirow{2}{*}[-0.8ex]{\textbf{Auxiliary Retriever}} &
            \multicolumn{2}{c}{\textbf{RULER @ 1M}} &
            \multicolumn{2}{c}{\textbf{LongBench}} \\
            \cmidrule(lr){3-4}
            \cmidrule(lr){5-6}
            & & GDN & Mamba-2 & GDN & Mamba-2 \\
            \midrule
            Random                  & \texttt{None} &  1.7 &  0.0 & 21.96 & 13.83 \\
            Input embedding         & \texttt{None} & 30.8 & 20.8 & 22.40 & \textbf{16.70} \\
            \rowcolor[HTML]{EAF1F8} \textbf{Native (NB)}
                                    & \texttt{None} & \textbf{59.8} & \textbf{37.6} & \textbf{23.03} & 14.25 \\
            \midrule
            BM25                    & \texttt{Lexical index}    & \textbf{64.0} & \textbf{52.7} & 23.50 & \textbf{18.66} \\
            BGE                     & \texttt{External encoder} & 45.4 & 34.0 & \textbf{23.77} & 17.24 \\
            \bottomrule
        \end{tabular}
        \captionof{table}{\textbf{Effect of routing strategy.} Scores average 6 RULER tasks at 1M tokens and 12 evaluated LongBench tasks, with LoRA readers fixed across routers for each backbone and benchmark. Native (NB) uses feature matching and voting; input embeddings are mean-pooled. Auxiliary retrievers add a lexical index or encoder; \texttt{None} does not imply zero routing cost.
        }
        \label{tab:router_compr:main}
    \end{minipage}
    \par
    \vspace{0.6em}
    \begin{minipage}{\textwidth}
        \centering
    
        \small
        \renewcommand{\arraystretch}{1.0}
        \setlength{\tabcolsep}{14pt}
        \begin{tabular}{lcccc}
            \toprule
            \multirow{2}{*}[-1ex]{\textbf{Memory Access}}
            & \multicolumn{2}{c}{\textbf{RULER @ 1M}}
            & \multicolumn{2}{c}{\textbf{LongBench}} \\
            \cmidrule(lr){2-3} \cmidrule(lr){4-5}
            & Frozen & LoRA
            & Frozen & LoRA \\
            \midrule
            \texttt{w/o} NoteBank        & 1.7           & 1.7           & 15.49          & 22.72          \\
            \texttt{w/ } NoteBank        & \textbf{59.8} & \textbf{58.4} & \textbf{16.07} & \textbf{23.03} \\
            \bottomrule
        \end{tabular}
        \vspace{-0.1em}
        \captionof{table}{\textbf{Separating memory access from read adaptation.}
        GDN scores with and without NoteBank, using frozen or LoRA-adapted readers.
        Scores average six RULER retrieval tasks at 1M tokens and twelve LongBench tasks.
        NoteBank uses top-1 replay; adapters are trained independently with and without replay using matched data and training budgets.
        }
    \label{tab:read_adaptation}
    \end{minipage}
\end{figure*}

\subsection{What does read adaptation contribute?}\label{sec:understand:adapter}
To distinguish evidence recovery from adaptation gains, we separate memory-access gains from adaptation gains using GDN with and without NoteBank, each with frozen or LoRA readers (Table~\ref{tab:read_adaptation}). Adapters are trained separately with and without replay using the same data and training budget; NoteBank uses top-1 retrieval. On RULER at 1M tokens, adaptation without NB leaves the score at 1.7, whereas NB-TF reaches 59.8. Adding LoRA yields 58.4, showing that this long-range retrieval recovery does not require read adaptation. On LongBench, adaptation improves scores from 15.49 to 22.72 without NB and from 16.07 to 23.03 with NB. The adapted models differ by only 0.31 points, indicating that adaptation accounts for most GDN's LongBench improvement in this setting. These controls distinguish two contributions: NoteBank enables long-range retrieval with frozen weights, while task adaptation drives most of the observed LongBench gains on GDN.

\section{Discussion and Limitations}\label{sec:discuss}
Our results challenge the view that linear attention must sacrifice long-context capability for efficiency, showing that separating recurrent computation from memorisation enables substantially stronger long-context performance at little additional cost. NoteBank realises this separation with sparse notes, native-feature addressing, and backbone replay. This \textbf{\textit{training-free, plug-and-play}} extension offers an \textbf{\emph{almost free lunch}}: 37.8-68.6 percentage-point gains at 1M-token retrieval across three linear attention architectures, with negligible marginal arithmetic, storage, and modest runtime overhead. Benefits extend to real-world tasks and a strong post-trained hybrid. This practical architecture leaves ample scope to improve writing, routing, and reading. Our interventions reveal opportunities to improve routing and evidence utilisation, with  gains varying across tasks and backbones. 
Training-free effectiveness establishes compatibility with existing models, while selectively training writing, routing, or reading may improve evidence preservation, access, and utilisation.

\newpage
\subsection*{AI use statement}
Generative AI tools were used to assist with code implementation and debugging, as well as language editing and presentation of the manuscript. The core research idea was developed by the authors. AI-assisted code was reviewed and tested by the authors before use in the experiments, and all AI-assisted manuscript content was manually reviewed and revised for technical correctness and consistency with the reported results. The authors take full responsibility for the final content of the paper, including all code, experiments, analyses, and claims produced with the assistance of generative AI.



\subsection*{Reproducibility statement}
We provide detailed information to support reproducibility in the appendix, including the complete NoteBank configuration, implementation details, benchmark and evaluation protocols, read-adaptation settings, and efficiency measurement procedures. We additionally report per-task and extended experimental results to facilitate verification of the main findings.


\subsection*{Acknowledgments}
The research reported in this publication was supported by funding from King Abdullah University of Science and Technology (KAUST) - Center of Excellence for Generative AI, under award number 5940.

\bibliography{refs}
\bibliographystyle{iclr2027/iclr2027_conference}

\clearpage
\appendix
\startcontents[appendix]


\section*{Appendix Contents}
\printcontents[appendix]{}{1}[2]{}

\bigskip

\section{Related Works}\label{sec:literature}
\paragraph{Linear and Recurrent Long-Context Models.} Recurrent sequence models~\citep{schmidhuber1992fwp,schmidhuber1992learning,hochreiter1997long,cho2014gru,beck2024xlstm} reduce the cost of quadratic self-attention by compressing the input prefix into a state whose size is independent of sequence length. Unlike conventional LSTMs with nonlinear recurrent dependencies, linear recurrent models use input-dependent linear state updates, enabling efficient parallel sequence processing while constraining how past state influences each update. Two representative families are selective state-space models, exemplified by Mamba and Mamba-2~\citep{gu2024mamba,dao2024transformers}, and delta-rule associative memories, exemplified by DeltaNet~\citep{schlag2021linear,yang2024deltanet} and Gated DeltaNet~\citep{yang2025gateddeltanet}. We study Mamba-2 and Gated DeltaNet variants to evaluate NoteBank across these distinct recurrent update mechanisms.
With fixed state size and numerical precision, a single pass cannot preserve arbitrarily growing histories losslessly. Reliable recall therefore depends on what the state retains and how stored associations interfere \citep{schlag2021linear,arora2024zoology,arora2024justreadtwice}. Hybrid architectures combine recurrent layers with attention to improve access to token-level representations \citep{lieber2024jamba,yang2024deltanet}. In particular, hybrid variants of Gated DeltaNet and Gated DeltaNet-2~\citep{yang2025gateddeltanet,hatamizadeh2026gateddeltanet2} incorporate sliding-window attention for local interactions and report improvements on evaluated retrieval tasks. In these sliding-window hybrids, direct attention access remains confined to the local window, while information from more distant inputs is carried through the recurrent computation. Across nonlinear and linear recurrent models, fixed-size states must accommodate an ever-growing history; adding attention provides local access but no persistent archive of distant evidence. 

\paragraph{Memory Extension for Recurrent Models.} 
Improving finite recurrent memory has long motivated designs such as LSTM gating~\citep{hochreiter1997long}. Linear recurrent models like Mambas~\citep{gu2024mamba,dao2024transformers} and Gated DeltaNet~\citep{yang2025gateddeltanet} simplify LSTM-style gating by computing gates from layer inputs independently of the recurrent state, retaining input-only memory control while enabling efficient parallel training. TTT layers~\citep{sun2024tttlayers} and Titans~\citep{behrouz2025titans} instead encode history through online learning, using self-supervised updates and momentum-based updates with forgetting, respectively. These approaches improve how memory is updated but retain fixed-size recurrent states, motivating extensions that preserve additional historical memory.
Memory Caching~\citep{behrouz2026memorycaching} additionally preserves multiple memory states, allowing effective memory capacity to grow with sequence length; it considers both continuous-recurrence checkpoints and independently initialized segment memories, and evaluates caching on Titans. MARCH~\citep{zhang2026march} caches cumulative recurrent-state anchors and uses content-dependent routing to combine historical readouts with the current-state output. HOLA~\citep{cui2026hola} augments the recurrent state with a bounded cache of key-value pairs selected according to the recurrent update magnitude. These caching methods differ in how much history they retain and whether they store compressed states or selected key-value pairs. Their retrieval operations read the retained states or values using current queries, without re-encoding the corresponding raw inputs. Appendix~\ref{app:res:complementary} compares against published LongBench results from Memory Caching and MARCH, whose evaluated models were trained on 16K-token sequences.


\paragraph{Retrieval and Replay.} Retrieval-based methods access external corpora or prior inputs and incorporate the retrieved information through representation fusion or renewed text processing. Retrieval-augmented generation (RAG) can provide textual evidence that need not have appeared in the current input history \citep{lewis2020rag}. Standard dense-retrieval pipelines select passages by matching query and passage embeddings and incorporate the retrieved text into the context for generation. Retrieval can also use the generation model's own representations: GritLM~\citep{muennighoff2024gritlm} unifies embedding and generation within a single language model, using a contrastive embedding objective to support corpus retrieval. Its corpus embeddings are produced through dedicated embedding passes, rather than extracted from the causal processing of the current input history. Another line of work revisits previously processed inputs. InfLLM~\citep{xiao2024infllm} extends pretrained Transformers without additional training by retrieving historical key-value blocks using representative token keys and incorporating them into attention. It reuses cached neural representations, whereas NoteBank retrieves archived text for native forward replay (see Figure~\ref{fig:efficiency_scaling}c and Table~\ref{tab:efficiency_1m} for direct comparison). ReadAgent~\citep{lee2024readagent} selects original pages through reasoning over textual gist summaries, while ReContext~\citep{zhao2026recontext} uses model-internal attention signals to select and replay evidence spans from the available prompt. Just Read Twice~\citep{arora2024justreadtwice} analyses how input order affects recall memory requirements and studies prompt repetition and non-causal prefix processing to exploit information unavailable during the initial causal pass.

Our approach stores raw historical tokens together with retrieval keys derived from model states during their initial causal processing. During subsequent causal computation, the current retrieval query selects tokens to replay through the model's input-processing path, allowing a new encoding after that query becomes available. 

\section{Algorithmic Details}\label{app:algo}
\begin{algorithm}[t]
\caption{\textbf{NoteBank}: Write, Address, and Replay}
\label{alg:notebank}
\small
\algrenewcommand{\algorithmicindent}{1em}
\algrenewcommand{\algorithmicrequire}{\textbf{Input:}}
\algrenewcommand{\algorithmicensure}{\textbf{Output:}}
\algnewcommand{\Hyperparameters}{\item[\textbf{HyperParams:}]}
\begin{algorithmic}[1]

\Require Frozen backbone $\mathcal{M}$; context $\mX$; question $\mQ$.
\Hyperparameters Chunk size $C$; write budget $W$; note length limit $B$; retrieval budget $R$.
\Ensure Generated response $\mY$.

\Statex
\Statex \textbf{1. Write}
    \hfill \textit{Before observing the question}
\State $\mathcal{S} \gets \mathcal{M}_{\phi}.\,\texttt{init\_state}$; \; 
    $\mathcal{NB} \gets [\,]$
\For{each chunk $\mX_{\mathcal{C}}$ in $\texttt{chunks}({\mX,C})$}
    \State $\mathcal{S}, \{\vk_{\mathcal{C}}\}, \{\vq_{\mathcal{C}}\} \gets
        \mathcal{M}_{\phi}.\,\texttt{forward}(\mathcal{S},\mX_{\mathcal{C}})$
    
    \State $\{d_{\mathcal{C}}\} \gets \Call{DistinctivenessMeasure}{\{\vk_{\mathcal{C}}\}}$ 
    \Comment{Eqn.~\ref{eq:nb-write}}
    
    \State $\mathcal{I} \gets
        \Call{SelectSpans}{\mX_{\mathcal{C}},\{d_{\mathcal{C}}\},W,B}$
    
    \For{each segment $I \in \mathcal{I}$}
        \State $\mathbf{z} \gets \Call{TokenIDs}{I}$ 
            \Comment{$I$: token positions in the sequence}
        
        \State $\mathcal{A} \gets
            \Call{AssignAddress}{\{\vq_{\mathcal{C}}\}, \{d_{\mathcal{C}}\}, I}$
            \Comment{Eqn.~\ref{eq:nb-address}}
        \State $\mathcal{NB}$.\,\texttt{append}$(\mathbf{z},\mathcal{A})$
    \EndFor
\EndFor

\Statex
\Statex \textbf{2. Address}
    \hfill \textit{Process the question for retrieval}
\State $\mathcal{Z} \gets [\,]$
\If{$|\mathcal{NB}| > 0$}
    \State $\mathcal{S}_{\mathrm{retrieval}} \gets
        \texttt{copy}(\mathcal{S})$
    \State $\_,\_,\{\vu\} \gets
        \mathcal{M}_{\phi}.\,\texttt{forward}(\mathcal{S}_{\mathrm{retrieval}}, \mQ)$    
        \Comment{$\vu \gets \operatorname{compact}(\vq_{\mQ})$ omitted}
    \State $V \gets
        \Call{Address}{\{\vu\},\, \mathcal{NB}.\,\texttt{addresses}}$
        \Comment{Eqn.~\ref{eq:nb-route}}
    \State $\{\mathbf{z}_{\mathrm{evidence}}\} \gets
        \Call{Top-R}{\mathcal{NB},V}$
    \State $\mathcal{Z} \gets
        \texttt{concat}(\{\mathbf{z}_{\mathrm{evidence}}\}.\,\texttt{sorted()})$
        \Comment{$\{\mathbf{z}_{\mathrm{evidence}}\}$ sorted by input order}
    \State $\texttt{delete}(\mathcal{S}_{\mathrm{retrieval}})$
\EndIf

\Statex
\Statex \textbf{3. Read}
    \hfill \textit{Replay, then answer}
\State $\mathcal{S}_{\mathrm{read}} \gets \texttt{copy}(\mathcal{S})$
\State $\mR \gets \mathcal{Z}$
\State $\mathcal{S}_{\mR} \gets
    \mathcal{M}_{\phi}.\,\texttt{forward}(\mathcal{S}_{\mathrm{read}}, \mR)$
    \Comment{Eqn.~\ref{eq:nb-replay}}
\State $Y \gets
    \mathcal{M}_{\phi}.\,\texttt{generate}(\mathcal{S}_{\mR}, \mQ)$
    \Comment{Eqn.~\ref{eq:nb-replay}}
\State \Return $Y$

\end{algorithmic}
\end{algorithm}
\subsection{Algorithmic overview} \label{sec:algorithm-overview}
Algorithm~\ref{alg:notebank} summarises the construction and use of NoteBank. During context processing, the writer maintains the
backbone state and builds an archive independently of the subsequent question. At query time, retrieval and readout branch from separate copies of the retained state: the question is first processed to select notes, after which the retrieved text and the question are processed on the answer branch. For clarity, the algorithm abstracts segment (or interchangeably, ``span'') selection, address construction, and matching into named operations. Chunk-boundary handling and sentence-level selection rules are specified in Appendix~\ref{app:write-details}.

\subsection{Write operation details} \label{app:write-details}
The writer operates at completed chunk boundaries, without observing the subsequent question. Within each chunk $\mathcal{C}_i$, we first $\ell_2$-normalise each native key, average the resulting unit vectors, and normalise the mean:
\begin{equation}
    \overline{\bm{\mu}}_i^{\ell,h} =
    \widehat{
        \frac{1}{|\mathcal{C}_i|}
        \sum_{t\in\mathcal{C}_i}
        \widehat{\vk}_t^{\ell,h}
    }.
\end{equation}
We average distinctiveness scores across layer-head channels to select text segments, and retain the per-channel scores to construct their addresses. To capture evidence that spans a chunk boundary, we extend the candidate region with the last 32 tokens of the preceding chunk. These tokens keep the scores and query features computed in their original chunk; their scores are not recomputed for the extended region. By default, we use this 32-token overlap and a 48-token scoring window.

Within each candidate region, the writer scores sliding windows by summing their token distinctiveness and selects the highest-scoring window. If the region is shorter than the configured window length, the window covers the entire region. We split the observed text at sentence-final punctuation and line breaks, and use the sentence containing the selected window’s central token as the initial span. We then expand the span by checking the immediately adjacent sentences, first on the left and then on the right. Each sentence is included only if the expanded span remains within the $B$-token budget. This process repeats until neither side can be added. Expansion follows sentence order and the length budget, without further salience comparisons. If the initial sentence alone exceeds $B$ tokens, we retain a contiguous $B$-token portion starting at the selected window’s start, adjusted to the nearest valid position that keeps the portion entirely within the sentence.

When writing two notes per chunk, we select the first note and exclude all sentences that overlap it. If the note contains only part of a long sentence, we exclude that entire sentence. This leaves at most two candidate regions, one on each side of the excluded text. We find the highest-scoring window across these regions and construct the second note using the same procedure. If no candidate remains, we write only one note. The two notes from a single write operation therefore do not overlap, although notes written at different chunk boundaries may overlap. For each note, we use the retained per-channel distinctiveness scores to select the query positions that form its addresses, as defined in Eqn.~\ref{eq:nb-address}. Our implementation supports $W\in\{1,2\}$. Any remaining tokens in a final incomplete chunk are processed by the backbone but produce no notes. A context shorter than $C$ tokens therefore leaves the archive empty.

\subsection{Native address construction} \label{app:address-construction}
For each selected note, we first identify one representative token per recurrent channel using the native-key distinctiveness scores,
as defined in Eqn.~\ref{eq:nb-address}. We then compact the native query at that position. Thus, representative-token selection uses key distinctiveness, while the stored address consists of query features.

We implement $\operatorname{compact}$ as fixed coordinate selection. For each channel $(\ell,h)$, we sample an ordered set of $d_\rho$ distinct coordinates from the native query dimension $d_k$, independently across channels, and keep this selection fixed across documents and questions. Let $P^{\ell,h}\in\{0,1\}^{d_\rho\times d_k}$ denote the corresponding coordinate-selection matrix. The note and question representations are
\begin{equation}
    \bm{\rho}_i^{\ell,h} = \operatorname{compact}\left( \vq_{t_i^{*,\ell,h}}^{\ell,h} \right)
    = P^{\ell,h}\vq_{t_i^{*,\ell,h}}^{\ell,h},
    \qquad
    \vu_\tau^{\ell,h} = \operatorname{compact}\left( \vq_\tau^{\ell,h} \right) = P^{\ell,h}\vq_\tau^{\ell,h}.
\end{equation}
Each row of $P^{\ell,h}$ selects one coordinate, with no coordinate repeated within a channel. The same mapping is used for stored notes and incoming questions; it introduces no trainable parameters. Our standard configuration uses $d_\rho=16$ and a fixed set of randomly-generated coordinates. Projected note addresses are stored in \texttt{BF16} without normalisation. The question representations are also converted to \texttt{BF16}. During matching, both are cast to \texttt{FP32} and $\ell_2$-normalised before computing cosine similarities. Compaction does not change which representative position is selected: that decision is made from the per-channel native-key distinctiveness scores before address compression.

\subsection{Replay readout and optional adaptation} \label{app:readout-adaptation}
The retained inference state $\mathcal{S}_T$ includes recurrent states, convolution caches, and, for hybrid backbones, attention key-value caches and position information. Retrieval and answer generation use separate copies of this state. As shown in Algorithm.~\ref{alg:notebank}, the question is first processed on a temporary copy to select notes; this copy is then discarded. On the answer branch, the selected notes are concatenated in document order and replayed before the question is processed again. Replay uses the original token IDs and follows the backbone's configured attention policy. Position indices continue from the context boundary rather than returning to the notes' original positions. An empty archive produces zero replay tokens.

Optional adaptation adds low-rank updates $\Delta W=(\alpha/r)BA$ to selected backbone projections, while keeping pretrained weights frozen. Here $r$ is the adapter rank and $\alpha$ controls its scale; $B$ is initialised to zero. For GDN backbones, adapters modify the query, key, value, and output projections. For Mamba-2, they modify the $x$, $B$, and $C$ projection branches
and the output projection. Adapters are disabled during context processing and retrieval, and enabled during replay, question processing, and answer generation.

We train the adapters using the mean negative log-likelihood of the target answer $\widetilde{\mY}$, including its EOS token: 
\begin{equation}
    \mathcal{L}
    = -\frac{1}{|\widetilde{\mY}|}
      \sum_{t=1}^{|\widetilde{\mY}|}
      \log p_{\phi,\theta}
      \bigl(\widetilde{\vy}_t
      \mid \mathcal{S}_{\mathrm{read}},\mR,\mQ,
      \widetilde{\vy}_{<t}\bigr),
\end{equation}
where $\phi$ denotes the frozen backbone weights and $\theta$ the adapter parameters. Only answer tokens contribute to the loss. The context state and discrete note selection are treated as constants, while gradients propagate through replay and subsequent question answering to the adapters.

\section{Experimental Details}\label{app:impl}
\subsection{Backbones} \label{app:backbones}
We evaluate NoteBank on GDN, GDN-2, Mamba-2, Mamba-2-8B, and Qwen3.5-9B. For the three smaller recurrent backbones, we use \texttt{m-a-p/1.3B-100B-GatedDeltaNet-pure}, \texttt{LLM-OS-Models2/gdn2-1.3b-paper-matched}, and \texttt{state-spaces/mamba2-1.3b}, respectively. GDN and GDN-2 contain 24 and 18 recurrent layers, yielding 192 and 288 indexed layer-head channels. Mamba-2 contains 48 layers with one B/C group per layer, yielding 48 independent addressing channels. GDN-2 and Mamba-2 checkpoints are converted to the local inference implementations, using their corresponding tokenizers. For Mamba-2-8B, we use NVIDIA's
\texttt{nvidia/mamba2-8b-3t-4k}, an 8B-parameter pure Mamba-2 base model pretrained on 3.5T tokens with a sequence length of 4K. Qwen3.5-9B (\texttt{Qwen/Qwen3.5-9B}) combines 24 Gated DeltaNet blocks with 8 Transformer self-attention blocks, giving a recurrent-to-attention ratio of 3:1. NoteBank uses native key and query features exclusively from the Gated DeltaNet blocks for writing and addressing, yielding 384 indexed layer-head channels. Replayed tokens pass through the entire hybrid backbone, including both block types, while the attention blocks retain their configured attention policy and KV caching. Pretrained backbone weights remain frozen in all NoteBank experiments; adapted variants train only the additional adapters. The selected backbones in summary:
\begin{itemize}
    \item GDN (1.3B):
        \href{https://huggingface.co/m-a-p/1.3B-100B-GatedDeltaNet-pure}
        {\texttt{m-a-p/1.3B-100B-GatedDeltaNet-pure}}.
    \item GDN-2 (1.3B):
        \href{https://huggingface.co/LLM-OS-Models2/gdn2-1.3b-paper-matched}
        {\texttt{LLM-OS-Models2/gdn2-1.3b-paper-matched}}.
    \item Mamba-2 (1.3B):
        \href{https://huggingface.co/state-spaces/mamba2-1.3b}
        {\texttt{state-spaces/mamba2-1.3b}}.
    \item Mamba-2 (8B, NVIDIA):
        \href{https://huggingface.co/nvidia/mamba2-8b-3t-4k}
        {\texttt{nvidia/mamba2-8b-3t-4k}}.
    \item Qwen3.5-9B:
        \href{https://huggingface.co/Qwen/Qwen3.5-9B}
        {\texttt{Qwen/Qwen3.5-9B}}.
\end{itemize}

\subsection{Benchmarks and evaluation protocols} \label{app:benchmarks}
\paragraph{RULER-NIAH~\citep{hsieh2024ruler}.}
We evaluate six retrieval tasks: \texttt{niah\_single\_1}, \texttt{niah\_single\_2}, \texttt{niah\_single\_3}, \texttt{niah\_multikey\_1}, \texttt{niah\_multiquery}, and \texttt{niah\_multivalue}. We use 100 fixed examples per task at each target length: 4K, 16K, 64K, 256K, and 1M tokens, where K denotes $2^{10}$. We apply the benchmark's task-specific scoring and report the unweighted mean across the six tasks at each length.

\paragraph{LongBench-v1~\citep{bai2024longbench}.}
We evaluate 12 subsets covering question answering, summarisation, few-shot tasks, and code completion: \texttt{narrativeqa}, \texttt{qasper}, \texttt{multifieldqa\_en}, \texttt{hotpotqa}, \texttt{2wikimqa}, \texttt{musique}, \texttt{gov\_report}, \texttt{qmsum}, \texttt{multi\_news}, \texttt{trec}, \texttt{triviaqa}, \texttt{samsum}. We use 30 fixed examples per subset (360 samples in total), shared across methods. The recurrent base models receive the native task prompts without additional chat formatting. We use the task-specific metrics and report their unweighted macro-average on a 0-100 scale.

\paragraph{MemoryAgentBench~\citep{hu2025memoryagentbench}.}
We evaluate Mamba-2-8B on 11 task configurations: FactConsolidation-sh and FactConsolidation-mh at 6K, 32K, and 262K; EventQA at 64K, 128K, and full length; and RULER QA1 and QA2. Each FactConsolidation and RULER configuration contains 100 evaluation questions, while each EventQA configuration contains 500, yielding 2,100 questions in total. The measured input lengths range from approximately 7K to 746K tokens. We compare plain inference, a clean-start baseline, and NoteBank with retrieval budgets $R=1$ and $R=4$. We report accuracy (\%) for each configuration and their unweighted macro-average.

\paragraph{LongBench-v2~\citep{bai2025longbenchv2}.}
For Qwen3.5-9B, we evaluate 375 full-input examples from \texttt{Single-Document QA}, \texttt{Multi-Document QA}, \texttt{Long-dialogue History Understanding}, and \texttt{User guide QA}. Examples are retained when the complete prompt fits within the position budget after reserving space for replay and generation; inputs are not truncated.
We use the official zero-shot prompt and answer parser, with the model's chat template and thinking disabled. Decoding is greedy with a maximum of 32,768 generated tokens, and the primary metric is accuracy over examples.
This evaluation covers the specified subset rather than the full benchmark.

\subsection{NoteBank and baseline configurations} \label{app:configurations}
For the short-note GDN, GDN-2, and Mamba-2 experiments, we use up to $W=2$ notes per completed chunk, a note budget of $B=64$ tokens, a 48-token salience window, and a 32-token rolling tail. The chunk size is $C=512$ for GDN and GDN-2 and $C=256$ for Mamba-2. Each note stores one $d_\rho=16$ dimensional address per channel, with random coordinate selection. We evaluate retrieval budgets $R\in\{1,2,4\}$. The plain baseline processes the original prompt without note replay; JRT-Prompt processes the context and question twice before generating an answer. Generation is greedy, with the same task-specific output limits and stopping rules across compared methods.

On very long inputs, we observe substantial recurrent-state magnitude explosion and degraded readout in Mamba-2 before any replay is performed. \textit{This instability already occurs in unmodified Mamba-2, independently of NoteBank or evidence replay}. Related work identifies state-distribution and dynamical mismatches as sources of length-generalisation failures~\citep{ruiz2025understanding,lu2025mambamodulation}. In our long-context setting, reliable readout therefore requires rescaling the recurrent state at the readout boundary $\mS^{\ell,h}\leftarrow\lambda\mS^{\ell,h}$ with a factor $\lambda\in[0,1]$, where $\lambda=1$ leaves the state unchanged and $\lambda=0.$ resets it. Small values of $\lambda$ often yield better performance on our longest-context tasks, both with and without NoteBank. 

For NoteBank, we simply use $\lambda=0.$ immediately before replay while retaining convolution caches and position information. Context prefill, note construction, and question-conditioned retrieval remain unchanged. Decoupling archival memory from recurrent computation makes this intervention possible: historical evidence remains accessible through retrieved text even after the accumulated recurrent state is discarded. Resetting alone also improves over unmodified Mamba-2 (Appendix~\ref{app:res:complementary}), suggesting that the accumulated state can impair subsequent readout in these settings.

We retain unmodified Mamba-2 as the main-text reference to report its out-of-the-box performance. To distinguish the effects of state rescaling from those of memory augmentation, this appendix additionally reports Mamba-2 baselines with the corresponding rescaling applied before question processing. NoteBank retains its advantage under these matched controls, showing that its gains are not explained by state rescaling alone.

\subsection{Adaptation setup} \label{app:adaptation-setup}
For each backbone and retrieval budget, we fine-tune the decoder on the same 2,400-example corpus using \emph{fewer than 9.3M tokens}, then evaluate the resulting adapter across all 12 LongBench tasks without task-specific fine-tuning. We construct this adaptation corpus from Natural Questions-derived short-answer examples in \texttt{emozilla/Long-Data-Collections-Fine-Tune}. Training contexts are capped at 16,000 characters, with gold-document identifier trailers removed and answers restricted to 1-24 words. Preserving the supplied data splits, the resulting pool contains 5,855 examples: 4,157 answerable and 1,698 natural no-support examples containing neither the answer string nor a gold-passage match. For both top-1 and top-2 adaptation, we use the same 2,400 examples without replacement after shuffling once with seed 0, comprising 1,722 answerable and 678 no-support examples. The total fine-tuning budget is below 9.3M tokens. No-support examples retain their original gold answers; supervised targets contain at most 47 answer tokens followed by EOS. A separate set of 354 no-support validation examples is used only for post-training diagnostics.

We use LoRA rank $r=8$, scale $\alpha=16$, and zero-initialised output factors at zero-indexed layers $\{11,15,18,22\}$ for GDN, $\{8,11,13,16\}$ for GDN-2, and $\{23,31,37,45\}$ for Mamba-2-1.3B. Training uses 600 AdamW updates with four examples accumulated per update, learning rate $5\times10^{-4}$, weight decay $0.01$, gradient clipping at $1.0$; evaluation uses the final checkpoint. Notes and routing are recomputed online with $C=512$, $W=2$, and $R\in\{1,2\}$. Adapter activation and the answer-only training objective follow App.~\ref{app:algo}.


\subsection{Implementation and efficiency measurement} \label{app:implementation-efficiency}
Efficiency measurements use an NVIDIA H100 80GB GPU, \texttt{BF16} precision, and batch size 1, with one benchmark process per GPU. For the GDN comparison, NoteBank uses $C=512$, $W=2$, $B=64$, and $R=2$. At each target length of 4K, 64K, and 1M, we measure three fixed RULER inputs: example 0 from \texttt{niah\_single\_2}, \texttt{niah\_multiquery}, and \texttt{niah\_multivalue}. Each input-method pair receives one warm-up run and three timed runs, with method order alternated. We report the mean of the per-input medians.

Timing covers complete prompt processing and generation of exactly 32 new tokens, with EOS and stop-string termination disabled. For NoteBank, this includes archive construction, question probing, routing, and replay; the archive is rebuilt for every run. Model loading, initial prompt tokenisation, warm-up, and result serialisation are excluded. Any additional tokenisation performed inside the NoteBank inference call remains included.

GPU memory is measured using \texttt{torch.cuda.max\_memory\_allocated()} and reported as an absolute peak including resident model weights and temporary allocations. Analytical persistent-storage estimates and host-side storage are reported separately. These fixed-output cost measurements are independent of the task-specific generation and scoring protocols used for quality evaluation.

For the Transformer comparisons, Full, SW4K, and InfLLM use the same checkpoint as in the main text, \texttt{m-a-p/transformer\_1.3B\_baseline}, with approximately 1.148B unique parameters, 24 layers, hidden width 2048, and 32 query/8 KV heads. Full uses FLA's causal FlashAttention and retains all historical KV across 512-token prefill chunks, whereas SW4K uses the same chunk size with a 4096-token causal window including the current token and no additional sinks. Both use unscaled RoPE with $\theta=10{,}000$ and monotonically increasing absolute positions; SW4K computes rotations on demand without altering this formula. InfLLM also uses 512-token chunks, processes the final prompt token separately, retains 128 initial tokens, and applies its official local/global positional treatment with the same RoPE base and unit distance scale. We apply no additional training or RoPE scaling: 1M inputs exceed the checkpoint's configured 4096-position range, although this configuration alone does not establish its original training sequence length.

\subsection{Added FLOPs Accounting} \label{app:added-flops}
We estimate the additional inference FLOPs of NB-TF relative to the same backbone without NoteBank, counting multiply-adds as two FLOPs. Accounting covers context ingestion through answer generation, using recorded input, archive, and replay lengths. Both methods use the corresponding Plain answer length $|\mY|$ to isolate method overhead. Let $J=|\mathcal L\times\mathcal H|$ and $|\mR|\leq RB$ be the replay length. 

\paragraph{Backbone cost.}
Let $f_\phi$ denote backbone FLOPs per token, excluding the language-model head, whose cost is $h_\phi=2DV$ for hidden width $D$ and vocabulary size $V$. We sum $2d_{\mathrm{in}}d_{\mathrm{out}}$ over projection and MLP matrices using checkpoint shapes, then add convolution and recurrent-equivalent costs $f_{\mathrm{conv}}$ and $f_{\mathrm{rec}}$.
For GDN,
\begin{equation}
\begin{aligned}
    f_{\mathrm{conv}}
        &=2c_{\mathrm{conv}}J(2d_k+d_v),\\
    f_{\mathrm{rec}}
        &=J(7d_kd_v+d_k+2d_v),
\end{aligned}
\end{equation}
where $c_{\mathrm{conv}}=4$.
The recurrent term covers state decay, key and query readouts, the rank-one update, and vector scaling. Our GDN configuration gives
$f_\phi\approx2.759\times10^9$ FLOPs/token and
$h_\phi\approx1.311\times10^8$ FLOPs/prediction.

\paragraph{NoteBank overhead.}
For a purely recurrent backbone,
\begin{equation}
    F_{\mathrm{Plain}}
    =(T+|\mQ|+|\mY|-1)f_\phi+|\mY|h_\phi.
\end{equation}
The final input position predicts the first answer token, requiring only $|\mY|-1$ subsequent backbone steps. With a nonempty archive,
\begin{equation}
\begin{aligned}
    \Delta F
    &\equiv F_{\mathrm{NB\text{-}TF}}-F_{\mathrm{Plain}}\\
    &=F_{\mathrm{write}}+F_{\mathrm{route}}
      +(|\mQ|+|\mR|)f_\phi.
\end{aligned}
\end{equation}
The additional backbone passes are the question match and replay; ordinary question processing and answer generation cancel. An empty archive incurs neither match nor routing or replay.

For the evaluated writer ($W=2$, with a 32-token tail),
\begin{equation}
\begin{aligned}
    F_{\mathrm{write}}\approx{}&
       6TJd_k+4\lceil T/C\rceil Jd_k+2TJ\\
       &+6(C+32)\lfloor T/C\rfloor.
\end{aligned}
\end{equation}
The first three terms estimate key normalisation, chunk-mean construction, distinctiveness scoring, and channel aggregation; the last conservatively bounds window-selection arithmetic. D addressing reuses these scores to select representative positions and gathers their compact query features, requiring no additional backbone pass or dense projection.

Matching every question token against all $N_T$ notes costs
\begin{equation}
    F_{\mathrm{route}}
    \approx2|\mQ|N_TJd_\rho
      +3(|\mQ|+N_T)Jd_\rho,
\end{equation}
including dot products and address normalisation.

\paragraph{Aggregation and interpretation.}
We calculate costs per example and report
\begin{equation}
    \mathrm{Added\ FLOPs ratio}
    =100\%\times
      \frac{\sum_e\Delta F_e}
           {\sum_eF_{\mathrm{Plain},e}}.
\end{equation}
For GDN with top-2 replay on 600 RULER examples at the 1M target length, mean Plain and additional costs are approximately $2891.4$ and $0.722$ TFLOPs, respectively, giving $0.025\%$ added FLOPs. Values are rounded after per-example accounting.

These estimates cover principal arithmetic, excluding most backbone pointwise operations, comparisons, sorting, indexing, data movement, and evaluator-only work. They use recurrent-equivalent costs rather than exact chunk-kernel execution counts and do not measure latency. The accounting excludes LoRA; hybrid backbones also require attention costs induced by replay, which do not cancel as above.

\subsection{Persistent Memory Accounting}
\label{app:persistent-memory}
We compare NoteBank's persistent state-and-archive payload with a full \texttt{BF16} MHA KV cache matched in hidden width $D$ and layer count $|\mathcal L|$. For input length $P=T+|\mQ|$, the reference occupies
\begin{equation}
    M_{\mathrm{MHA}}
    =4|\mathcal L|DP
    \quad\text{bytes},
\end{equation}
counting keys and values at two bytes per element, without KV-head sharing.

The GDN used in our experiments has 24 layers and 8 heads per layer, with $J=192$ and $d_k=d_v=256$, giving $D=2048$. NoteBank uses $C=512$, $W=2$, $B=64$, and $d_\rho=16$. Across 600 RULER examples at the 1M target length, the mean input length, note count, and archived-token count $U=\sum_{i=1}^{N_T}|\mathbf z_i|$ are
\begin{equation}
    \bar P\approx1{,}048{,}019,\qquad
    \overline{N_T}\approx4092.3,\qquad
    \bar U\approx232{,}514.
\end{equation}
We use these recorded counts rather than assuming
full-length notes or exactly $2^{20}$ input tokens.

Charging addresses as BF16, token IDs as compact
int32, and metadata as three uint64 fields per note
(\texttt{chunk}, \texttt{start}, \texttt{end}),
the archive payload is
\begin{equation}
    \mathrm{Storage}_{\mathcal{NB}}
    =2N_TJd_\rho+4U+24N_T
    \quad\text{bytes}.
\end{equation}
Including \texttt{FP32} recurrent states, \texttt{BF16} Q/K/V convolution caches, and resident helper tensors gives
\begin{equation}
\begin{aligned}
    M_{\mathrm{NB}}
    ={}&\mathrm{Storage}_{\mathcal{NB}}
       +4Jd_kd_v\\
       &+2c_{\mathrm{conv}}J(2d_k+d_v)
       +M_{\mathrm{helper}}.
\end{aligned}
\end{equation}
Here $c_{\mathrm{conv}}=4$ and $M_{\mathrm{helper}}\approx0.402$ MiB include projection indices and allocated auxiliary tensors, including those unused by routing. Using $1\,\mathrm{MiB}=2^{20}$ bytes, recurrent states, convolution caches, and the archive occupy 48, 1.125, and 24.959 MiB on average. Thus,
\begin{equation}
    \frac{\overline{M_{\mathrm{NB}}}}
         {\overline{M_{\mathrm{MHA}}}}
    \approx
    \frac{74.49}{196{,}503.5}
    \approx\frac{1}{2638}.
\end{equation}
This is a ratio of aggregate payloads, with rounding applied after accounting.

The comparison is taken at the start of answer generation and includes the entire archive. It excludes model weights, transient workspace, and encoding/replay cache copies. CPU text and metadata are charged at compact representation sizes, excluding Python-object overhead; these figures therefore describe payload, not measured CPU heap usage or peak GPU allocation. The ratio applies to this GDN configuration at 1M.

\section{Additional Results}\label{app:res}
\subsection{Complementary results}\label{app:res:complementary}
\begin{figure}[t]
    \begin{center}
        \includegraphics[width=\linewidth]{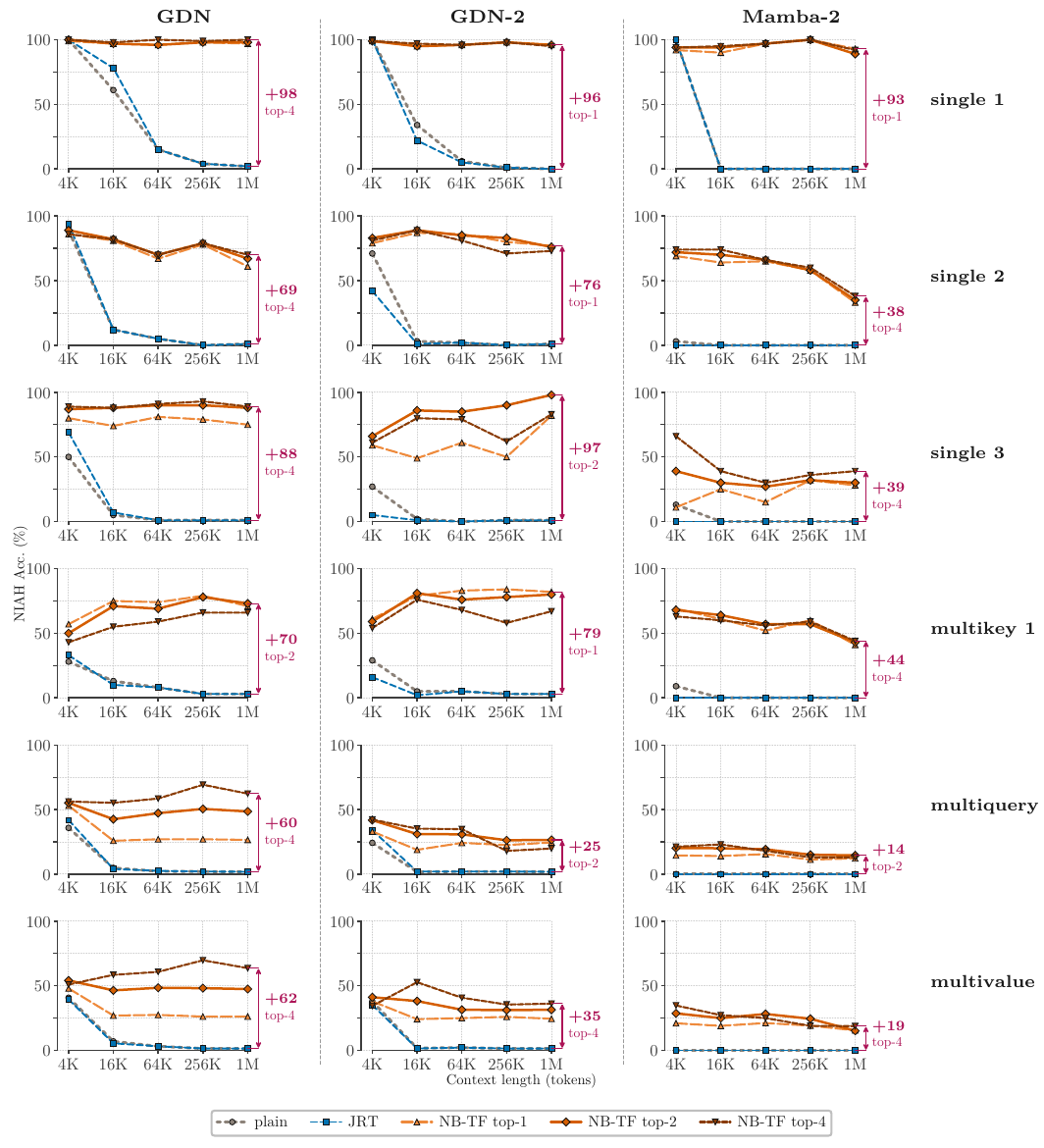}
    \end{center}
    \caption{\textbf{Training-free long-range retrieval across architectures and context lengths:} full results on RULER 6 NIAH tasks.}
    \label{fig:tf_ruler:full}
\end{figure}
\paragraph{Per-task retrieval results.}
Section~\ref{sec:recover:tf} reports accuracy averaged over six RULER tasks. Here, we provide the per-task breakdown for GDN, GDN-2, and Mamba-2 across context lengths from 4K to 1M, using the same checkpoints, methods (w/ and w/o NB-TF), and 100 examples per task. The breakdown shows that NoteBank's gains vary across task types: single-needle retrieval is generally stronger, while multi-query and multi-value tasks remain more challenging under a limited replay budget. NB-TF sustains substantial retrieval accuracy at long contexts where Plain and JRT largely fail. However, recovery is task-dependent: strong single-needle retrieval does not uniformly translate into successful multi-query or multi-value retrieval. The per-task breakdown also reveals greater variability for Mamba-2, which ties with only Mamba-2, showing that a shared memory interface does not eliminate backbone-specific limitations.

\begin{table*}[t]
    \centering
    \small
    \setlength{\tabcolsep}{3.5pt}
    \begin{tabular}{l|l|cccc|c}
            \toprule
            \textbf{Backbone} & \textbf{Method}
            & \textbf{Single-doc QA}
            & \textbf{Multi-doc QA}
            & \textbf{Summarisation}
            & \textbf{Few-shot}
            & \textbf{Overall} \\
            \midrule
            Transformer\textsuperscript{\dag}
            & Plain    & 13.40 & 18.97 & 8.43 & 23.40 & 16.05 \\
            \midrule
            Titans\textsuperscript{\dag}
            & SSC  & 12.80 & 18.07 & 7.63 & 27.50 & 16.50 \\
            GDN\textsuperscript{\ddag}
            & MARCH    &  8.87 & 7.90 & 13.23 & 29.47 & 14.87 \\
            \midrule
            & Plain    & 6.88 & 7.02 & 13.99 & 34.08 & 15.49 \\
            & JRT      & 7.72 & 6.63 & \textbf{14.79} & 33.56 & 15.67 \\
            \rowcolor[HTML]{EAF1F8}\cellcolor{white}
            & \textbf{NB-TF}    & 6.66 & 7.33 & 13.52 & 36.76 & 16.07 \\
            \rowcolor[HTML]{EAF1F8}\cellcolor{white}\multirow{-4}{*}{GDN}
            & \textbf{NB-LoRA} & \textbf{19.06} & \textbf{20.48} & 11.37 & \textbf{41.22} & \textbf{23.03} \\
            \midrule
            & Plain    & 2.22 & 2.10 & 11.27 & 11.56 & 6.79 \\
            & JRT      & 2.39 & 1.89 & 9.21  & 8.34  & 5.12 \\
            \rowcolor[HTML]{EAF1F8}\cellcolor{white}
            & \textbf{NB-TF}    & 5.98 & 5.36 & \textbf{16.51} & 25.88 & 13.43 \\
            \rowcolor[HTML]{EAF1F8}\cellcolor{white}\multirow{-4}{*}{Mamba-2}
            & \textbf{NB-LoRA} & \textbf{17.84} & \textbf{15.54} & 7.46 & \textbf{27.52} & \textbf{17.10} \\
            \bottomrule
        \end{tabular}
    \caption{\textbf{From retrieval to real-world long-context tasks.} 
    Scores are macro-averaged over evaluated LongBench-v1 tasks, within each category and overall. NB-TF requires no additional training; NB-LoRA adapts the read stage. \textsuperscript{\dag}Reference results aggregated from \citet{behrouz2026memorycaching}; \textsuperscript{\ddag}Reference results aggregated from \citet{zhang2026march}. Both \textsuperscript{\dag} and \textsuperscript{\ddag} require pretraining on 16K-context length data; out GDN and Mamba-2 were pretrained on 2K and 4K-length sequences, respectively. Bold marks the best score within each recurrent backbone.}
    \label{tab:longbench:full}
\end{table*}

\paragraph{Comparison with pretrained state-caching methods.}
Table~\ref{tab:longbench:full} extend Sec.~\ref{sec:recover:realworld} with published results for Memory Caching (MC; Titans+SSC)~\citep{behrouz2026memorycaching} and MARCH~\citep{zhang2026march}. Both incorporate state caching during pretraining and use 16K training sequences for their LongBench results; MARCH is pretrained on 50B tokens. These results provide contextual references, given differences in model sizes and pretraining protocols. Using an off-the-shelf GDN checkpoint pretrained on short contexts (\texttt{m-a-p/1.3B-100B-GatedDeltaNet-pure}, trained on $2K$), NB-TF achieves 16.07 overall, exceeding MARCH (14.87) and approaching MC (16.50), with \emph{no additional training}. NB-LoRA further reaches 23.03 using fewer than 9.3M adaptation tokens. Beyond LongBench, GDN with NB-TF retains approximately 70\% mean accuracy across six RULER tasks at 1M tokens (Sec.~\ref{sec:recover:tf}), without context-extension training. Together, these results demonstrate that effective long-range memory can be added to existing checkpoints without pretraining a specialised memory architecture.

\begin{table*}[t]
    \centering
    \small
    \setlength{\tabcolsep}{6pt}
    \begin{tabular}{l|l|cccc|c}
        \toprule
        \textbf{Backbone} & \textbf{Method}
        & \shortstack{\textbf{FactCon.}\\\textbf{Single-hop}}
        & \shortstack{\textbf{FactCons.}\\\textbf{Multi-hop}}
        & \textbf{EventQA}
        & \textbf{RULER QA}
        & \textbf{Overall} \\
        \midrule
        & Plain
        & 0.0 & 0.0 & 0.0 & 0.0 & 0.0 \\
        & Clean start
        & 14.0 & 2.0 & 17.7 & 16.0 & 13.0 \\
        \rowcolor[HTML]{EAF1F8}\cellcolor{white}
        & \textbf{NB-TF} (top-1)
        & 19.0 & \textbf{3.0} & \textbf{17.9}
        & 16.5 & 14.5 \\
        \rowcolor[HTML]{EAF1F8}\cellcolor{white}
        \multirow{-4}{*}{Mamba-2-8B}
        & \textbf{NB-TF} (top-4)
        & \textbf{22.0} & 2.5 & 16.4
        & \textbf{17.0} & \textbf{14.7} \\
        \midrule

        & Plain
        & 42.5 & \textbf{5.5} & 90.2 & 67.5 & 55.7 \\
        \rowcolor[HTML]{EAF1F8}\cellcolor{white}
        & \textbf{NB-TF}(top-1)
        & 42.5 & \textbf{5.5} & 90.2 & 67.5 & 55.7 \\
        \rowcolor[HTML]{EAF1F8}\cellcolor{white}
        \multirow{-3}{*}{Qwen3.5-9B}
        & \textbf{NB-TF}(top-4)
        & \textbf{45.0} & \textbf{5.5} & \textbf{90.7}
        & \textbf{69.0} & \textbf{56.8} \\
        \bottomrule
    \end{tabular}
    \caption{
        \textbf{MemoryAgentBench results by task family and replay budget.}
        Accuracy (\%) for Mamba-2-8B and Qwen3.5-9B,
        using matched examples within each backbone.
        RULER QA differs from the RULER-NIAH tasks reported in this paper.
        Overall reports the macro-average over evaluated tasks.
        For Mamba-2, Clean start resets the inference state
        without replay, while NB-TF additionally replays
        retrieved evidence.
        Bold marks the best score within each backbone,
        including ties.
    }
    \label{tab:memoryagentbench:full}
\end{table*}
\paragraph{Task-family results and replay budgets on MemoryAgentBench.}
Table~\ref{tab:memoryagentbench:full} expands Section~\ref{sec:recover:plug-and-play} with task-family results at $R=1$ and $R=4$. For Qwen3.5-9B, top-1 matches Plain, while top-4 increases overall accuracy from 55.7 to 56.8. Larger replay budgets are not uniformly beneficial: for Mamba-2-8B, increasing $R$ improves single-hop FactConsolidation from 19.0 to 22.0 but reduces EventQA accuracy from 17.9 to 16.4.

\paragraph{State reset and evidence replay in Mamba-2.}
As discussed in App.~\ref{app:configurations}, Mamba-2 suffers severely from long-context state explosion, and by decoupling memorisation and recurrent computation, NoteBank facilitate a bypass: resetting the recurrent computation when performing QA and answering questions using only the retrieved evidence. Table~\ref{tab:memoryagentbench:full}, shows that clean start raises Mamba-2-8B's overall accuracy from 0.0 to 13.0, showing that the accumulated inference state substantially impairs readout. NB-TF further improves accuracy to 14.5 with top-1 and 14.7 with top-4 replay. The largest gain over Clean start occurs on single-hop FactConsolidation, from 14.0 to 22.0. Thus, resetting accounts for much of the recovery, while retrieved evidence provides additional, task-dependent benefits.

\begin{figure}[t]
    \centering
    \includegraphics[width=\linewidth]{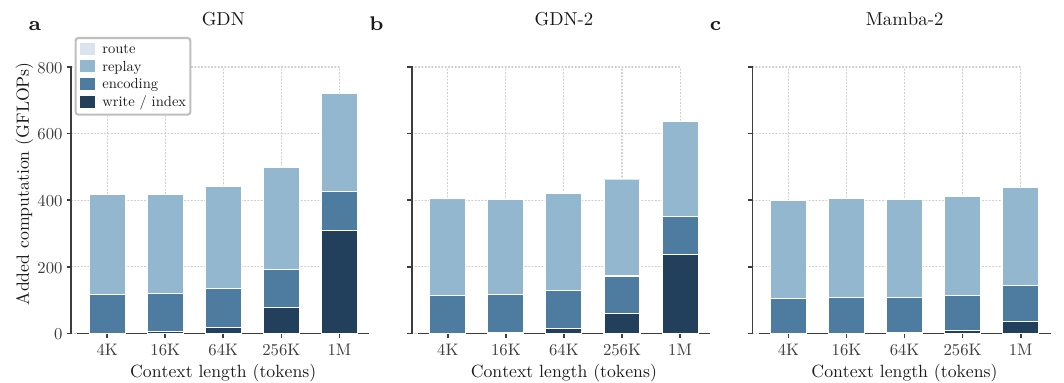}
    \caption{
        \textbf{Breakdown of NoteBank's computation overhead.}
        Added analytical FLOPs for NB-TF with top-2 replay
        relative to Plain on RULER, using matched output lengths.
        Stacked bars separate writing/indexing, question encoding,
        routing, and replay for GDN, GDN-2, and Mamba-2. 
        Annotations report routing costs at 1M tokens in GFLOPs.
    }
    \label{fig:efficiency:breakdown}
\end{figure}

\paragraph{Efficiency breakdown.}
Figure~\ref{fig:efficiency:breakdown} decomposes the
additional analytical FLOPs of NB-TF into writing/indexing, question encoding, routing, and replay. Writing and indexing grow linearly with context length but reuse native features without another backbone pass. Question encoding and replay remain approximately constant for fixed question length and replay budget. Routing grows with the archive size, yet matching compact addresses costs only 1.09, 1.72, and 0.50 GFLOPs at 1M tokens for GDN, GDN-2, and Mamba-2, respectively. Thus, operations that scan the context or archive remain lightweight, while additional backbone computation is confined to the question and retrieved evidence. Together, these costs add only 0.0171\%--0.0250\% to full-model analytical FLOPs at 1M tokens. These arithmetic estimates do not imply an equivalent latency overhead.

\paragraph{Practical runtime and memory discussion}
\begin{table*}[t]
    \centering
    \small
    \setlength{\tabcolsep}{14pt}
    \begin{tabular}{lrrrrr}
        \toprule
        & \multicolumn{2}{c}{\textbf{Gated DeltaNet}}
        & \multicolumn{3}{c}{\textbf{Transformer}} \\
        \cmidrule(lr){2-3} \cmidrule(lr){4-6}
        \textbf{Metric}
        & Plain & \textbf{+ NB-TF}
        & Full & + SW4K & + InfLLM \\
        \midrule
        RULER @ 1M $\uparrow$               & 1.71          & \textbf{70.33} & 0  & 1.17 & 0     \\
        GPU peak (GiB) $\downarrow$    & \textbf{2.93} & 3.27           & 55.18 & 5.41 & 7.30     \\
        CPU archive (MiB) $\downarrow$ & 0             & 0.98           & 0     & 0    & 48930.00  \\
        \bottomrule
    \end{tabular}
    \caption{
    \textbf{Retrieval quality vs. measured storage at 1M tokens.}
        RULER averages six NIAH tasks.
        GPU peak includes model weights.
        CPU archive reports compact payload accounting for NB's token IDs and span-location records for NoteBank, and counted host KV for InfLLM.
    }
    \label{tab:efficiency_1m}
\end{table*}
Table~\ref{tab:efficiency_1m} relates retrieval quality to GPU and CPU memory footprints at 1M tokens. The three Transformer configurations share the same checkpoint. Full attention requires 55.18\,GiB of peak GPU memory. InfLLM uses 7.30\,GiB of peak GPU memory while retaining 47.78\,GiB of historical KV on the CPU. SW4K achieves lower latency at 61.38\,s, but scores only 1.17 on RULER, compared with 70.33 for GDN with NoteBank. These cross-backbone comparisons characterise the evaluated systems, whereas the paired GDN measurements isolate NoteBank's incremental cost. GPU figures report measured peak allocated memory, including model weights; CPU archive figures report compact payload accounting for NoteBank and counted host KV bytes for InfLLM.


\paragraph{Retrieval: quality vs cost.}
\begin{table*}[t]
    \centering
    \renewcommand{\arraystretch}{1.0}
    \setlength{\tabcolsep}{2.8pt}
    \begin{tabular}{llccccc}
        \toprule
        \multirow{2}{*}[-0.8ex]{\textbf{Routing Signal}} &
        \multirow{2}{*}[-0.8ex]{\textbf{Auxiliary Retriever}} &
        \multirow{2}{*}[-0.8ex]{\textbf{Time @ 1M (ms)}} &
        \multicolumn{2}{c}{\textbf{RULER @ 1M}} &
        \multicolumn{2}{c}{\textbf{LongBench}} \\
        \cmidrule(lr){4-5}
        \cmidrule(lr){6-7}
        & & & GDN & Mamba-2 & GDN & Mamba-2 \\
        \midrule
        Random          & \texttt{None}             & --    &  1.7 &  0.0 & 21.96 & 13.83 \\
        Input embedding & \texttt{None}             & 7,600 & 30.8 & 20.8 & 22.40 & \textbf{16.70} \\
        \rowcolor[HTML]{EAF1F8}
        \textbf{Native (NB)}
                        & \texttt{None}             & 26    & \textbf{59.8} & \textbf{37.6} & \textbf{23.03} & 14.25 \\
        \midrule
        BM25            & \texttt{Lexical index}    & 87    & \textbf{64.0} & \textbf{52.7} & 23.50 & \textbf{18.66} \\
        BGE             & \texttt{External encoder} & 2,630 & 45.4 & 34.0 & \textbf{23.77} & 17.24 \\
        \bottomrule
    \end{tabular}
    \caption{\textbf{Routing quality and computational cost.}
    Scores average six RULER tasks at 1M tokens and 12
    LongBench tasks, with LoRA readers fixed across routers
    within each backbone and benchmark.
    Bold marks the best score within each auxiliary-retriever
    group. Native routing matches backbone features and
    aggregates channel votes; input embeddings are mean-pooled.
    Reported times use a 4092-note bank at 1M:
    Native includes question probing and voting,
    BM25 runs on CPU, and BGE reports full-bank encoding.
    Timing scopes differ and do not constitute a matched
    query-latency comparison; no Random timing is available.
    \texttt{None} indicates no auxiliary retriever,
    not zero computational cost.}
    \label{tab:router_compr:full}
\end{table*}
Table~\ref{tab:router_compr:full} supplements the main-text quality comparison with runtime costs for a 4092-note bank at 1M tokens. For GDN, native routing takes 26\,ms, including question probing, feature matching, and channel voting. CPU BM25 takes 87\,ms in our implementation and achieves the highest RULER scores. This illustrates an alternative quality-cost trade-off within the same archive and reader. The input-embedding and BGE entries report 7,600 and 2,630\,ms, respectively, but their timing scopes differ; in particular, BGE includes full-bank encoding. These figures therefore describe the reported implementation costs rather than a matched per-query latency benchmark. The small measured overhead of the native query-and-route path, together with its retrieval quality and lack of an auxiliary encoder or lexical index, supports native routing as a practical default while leaving room for alternative routers to improve quality.

\subsection{Ablation study}\label{app:res:ablation}
We study the per-chunk write budget $W$, replay budget $R$, span budget $B$, and chunk size $C$ on GDN and Mamba-2. Training-free ablations report mean accuracy across six RULER tasks from 4K to 1M tokens, together with analytical added FLOPs where applicable. We additionally examine the interaction between $R$ and $B$ with NB-LoRA on twelve LongBench tasks.

\paragraph{Write budget $W$.}
\begin{figure}[t]
    \centering
    \includegraphics[width=\linewidth]{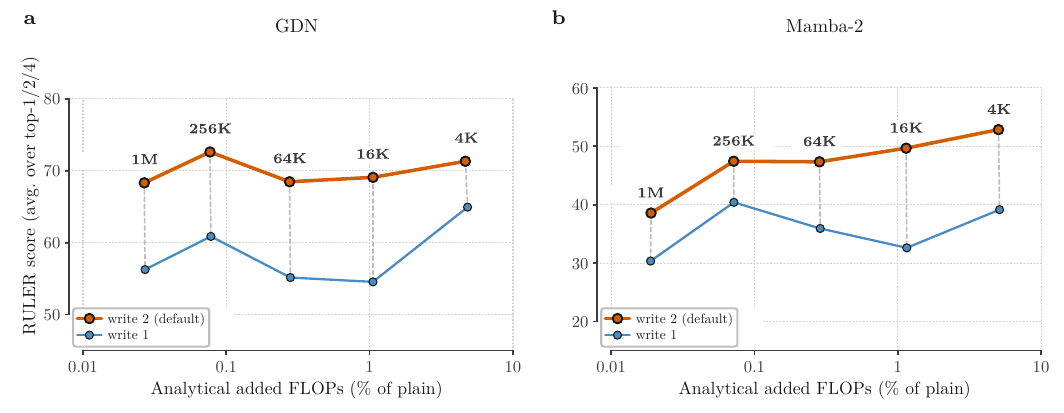}
    \caption{
        \textbf{Effect of the per-chunk write budget.}
        Six-task mean RULER accuracy versus analytical added FLOPs for NB-TF with $W\in\{1,2\}$ on GDN and Mamba-2, averaged over replay budgets $R\in\{1,2,4\}$. Labels indicate context lengths. Writing two notes consistently improves accuracy with little additional arithmetic cost.
    }
    \label{fig:ablation:write}
\end{figure}
We test whether retaining more notes per chunk improves retrieval by comparing $W=1$ and $W=2$, averaging over $R\in\{1,2,4\}$ (Figure~\ref{fig:ablation:write}). Writing two notes consistently improves accuracy across both backbones and all evaluated context lengths. The gain comes with little change in analytical added FLOPs, since native feature extraction and distinctiveness scoring are shared across notes. This supports $W=2$ as an effective way to improve archive coverage, at the cost of storing more notes.

\paragraph{Replay budget $R$.}
\begin{figure}[t]
    \centering
    \includegraphics[width=\linewidth]{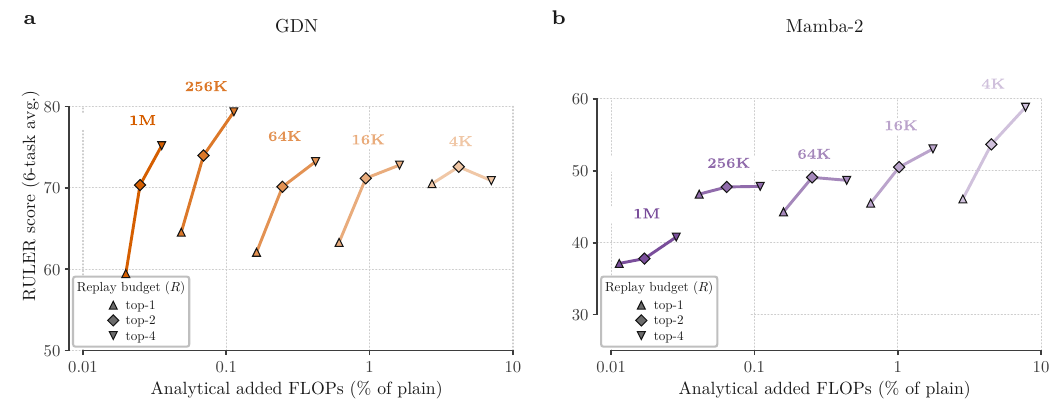}
    \caption{
        \textbf{Effect of the replay budget on retrieval and computation.}
        Six-task mean RULER accuracy versus analytical
        added FLOPs for NB-TF with $R\in\{1,2,4\}$ on GDN and Mamba-2.
        Labels indicate context lengths; connected markers compare replay budgets at the same length.
        Additional replay generally improves accuracy, with gains depending on the backbone and context length.
    }
    \label{fig:ablation:replay}
\end{figure}
We vary $R\in\{1,2,4\}$ to assess the benefit and computational cost of replaying additional evidence (Figure~\ref{fig:ablation:replay}). At 1M tokens, increasing $R$ from 1 to 4 raises GDN accuracy from 59.5\% to 75.2\% and Mamba-2 accuracy
from 37.1\% to 40.8\%. The benefit is not uniform: GDN peaks at $R=2$ at 4K,
and Mamba-2 shows small or negligible gains in some long-context settings. Larger budgets increase replay computation, but its relative cost decreases as context ingestion accounts for more of the full-model FLOPs. Thus, additional replay can improve retrieval, with diminishing returns that depend on the backbone and context length.

\paragraph{Chunk size $C$.}
\begin{figure}[t]
    \centering
    \includegraphics[width=\linewidth]{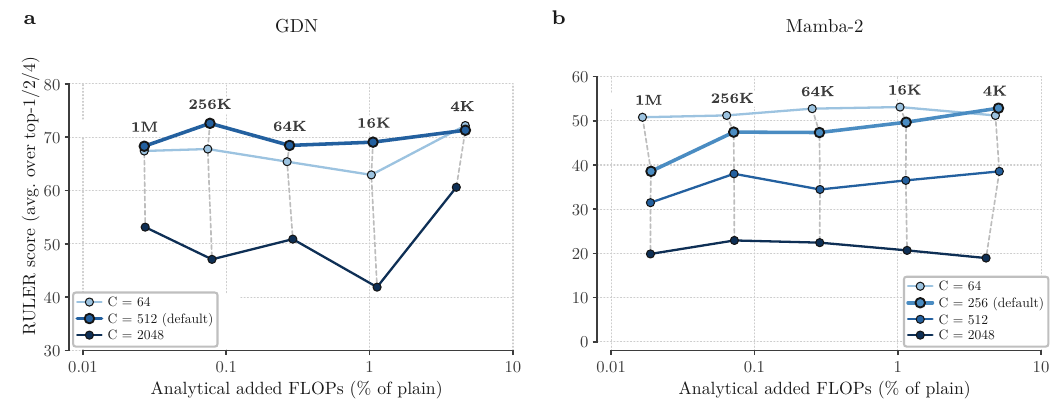}
    \caption{
        \textbf{Effect of chunk size on retrieval and computation.}
        Six-task mean RULER accuracy versus analytical added FLOPs for NB-TF with different chunk sizes $C$, averaged over replay budgets $R\in\{1,2,4\}$.
        Labels indicate context lengths.
        Smaller chunks particularly benefit Mamba-2 at long contexts, while very large chunks degrade retrieval for both backbones.
    }
    \label{fig:ablation:chunk}
\end{figure}
We vary $C$ from 64 to 2048 to examine how writing granularity affects retrieval, averaging over $R\in\{1,2,4\}$ (Figure~\ref{fig:ablation:chunk}). GDN's default $C=512$ matches or exceeds $C=64$ across the evaluated lengths, whereas $C=2048$ substantially reduces accuracy. Mamba-2 benefits more from frequent writing: $C=64$ outperforms its default $C=256$ at long contexts, while $C=512$ and $2048$ perform worse. The FLOPs differences are small, although smaller chunks permit denser archives and therefore increase storage requirements. These results show that useful writing granularity depends on the backbone, and that very coarse chunks can discard opportunities to preserve evidence.

\paragraph{Span budget $B$.}
\begin{figure}[t]
    \centering
    \includegraphics[width=\linewidth]{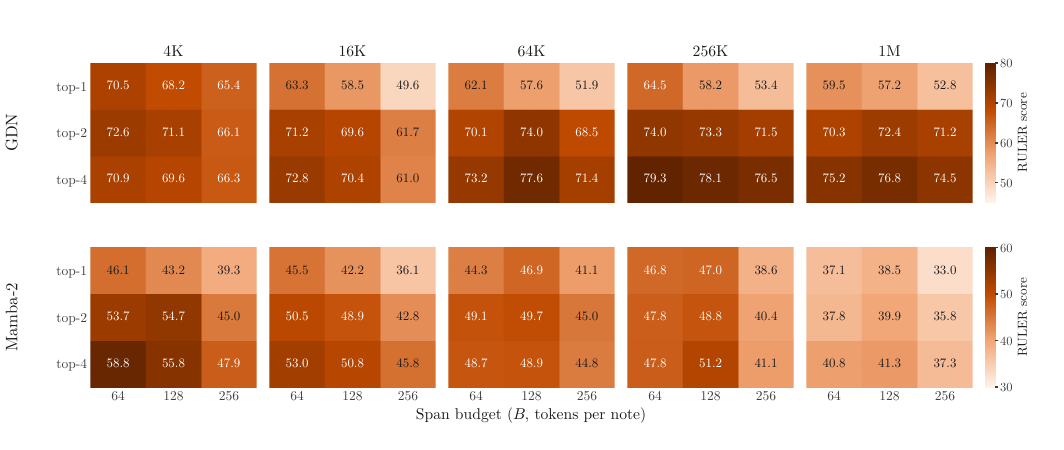}
    \caption{
        \textbf{Joint effects of span and replay budgets.}
        Six-task mean RULER accuracy (\%) for NB-TF on
        GDN (top) and Mamba-2 (bottom), at context lengths
        from 4K to 1M.
        Each heatmap varies the span budget
        $B\in\{64,128,256\}$ across columns and replay budget
        $R\in\{1,2,4\}$ across rows.
        Longer spans do not consistently improve accuracy;
        their benefit depends on the replay budget and backbone.
    }
    \label{fig:ablation:span}
\end{figure}
We jointly vary $B\in\{64,128,256\}$ and $R\in\{1,2,4\}$ to distinguish longer notes from replaying more notes (Figure~\ref{fig:ablation:span}). For GDN, $B=64$ performs best with top-1 replay at every evaluated length, while $B=128$ often helps
with larger replay budgets at long contexts. For Mamba-2, $B=256$ consistently underperforms both smaller span budgets at matched replay budgets. At 1M, the best settings in this grid use $(B,R)=(128,4)$, reaching 76.8\% for GDN and 41.3\% for Mamba-2. Larger text budgets therefore do not automatically improve retrieval; span length and evidence coverage must be considered jointly.


\paragraph{Interaction: replay $\times$ span budget.}
\begin{figure}[t]
    \centering
    \includegraphics[width=\linewidth]{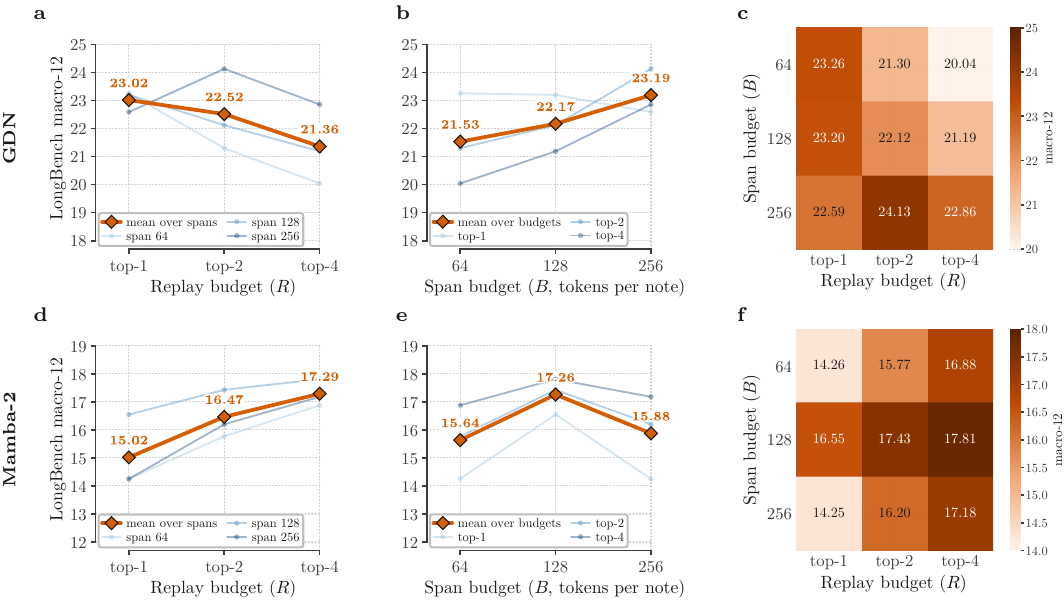}
    \caption{
        \textbf{Replay--span interaction with read adaptation.}
        LongBench macro-average scores over twelve evaluated
        tasks for NB-LoRA on GDN (top) and Mamba-2 (bottom).
        Left: varying replay budget $R$, with averages over
        span budgets. Middle: varying span budget $B$, with
        averages over replay budgets. Right: scores for each
        $(B,R)$ configuration.
        The best replay and span budgets depend on the backbone.
    }
    \label{fig:ablation:longbench}
\end{figure}
We evaluate the same $R$-$B$ grid with NB-LoRA on LongBench, reporting marginal averages and the full configuration heatmaps (Figure~\ref{fig:ablation:longbench}).
Averaged over span budgets, increasing $R$ from 1 to 4 reduces GDN's score from 23.02 to 21.36. However, the heatmaps reveal an interaction hidden by these averages: GDN performs best at $(B,R)=(256,2)$ with a score of 24.13, while Mamba-2 peaks at 17.81 with $(B,R)=(128,4)$. For Mamba-2, $B=128$ is also the best span budget at every replay budget. Read adaptation therefore does not remove sensitivity to evidence quantity and granularity, and the best configuration remains backbone-dependent.

\stopcontents[appendix]

\end{document}